\documentclass[lettersize,journal]{IEEEtran}
\usepackage{amsmath,amsfonts}
\usepackage{array}
\usepackage[caption=false,font=normalsize,labelfont=sf,textfont=sf]{subfig}
\usepackage{textcomp}
\usepackage{stfloats}
\usepackage{url}
\usepackage{verbatim}
\usepackage{graphicx}

\usepackage[numbers]{natbib}
\usepackage{algorithm}
\usepackage{algpseudocode}

\usepackage{bm}
\usepackage{xcolor}

\usepackage{booktabs}
\usepackage{amssymb}

\DeclareMathAlphabet{\pazocal}{OMS}{zplm}{m}{n}

\usepackage{mathtools,xparse}
\DeclarePairedDelimiter{\rightbarinner}{.}{\rvert}
\NewDocumentCommand{\rightbar}{som}{%
  \IfBooleanTF{#1}
    {\rightbarinner*{#3}}
    {\IfNoValueTF{#2}{#3\rvert}{\rightbarinner[#2]{#3}}}%
}

\usepackage[switch]{lineno}

\begin{document}

\title{Novel and flexible parameter estimation methods for data-consistent inversion in mechanistic modeling}

\author{
    Timothy Rumbell\textsuperscript{\rm 1},
    Jaimit Parikh\textsuperscript{\rm 1}, 
    James Kozloski\textsuperscript{\rm 1},
    Viatcheslav Gurev\textsuperscript{\rm 1*}
\thanks{
        \textsuperscript{\rm 1} IBM Research, Hybrid Biological-AI Modeling
        
        \textsuperscript{\rm *} Corresponding author: vgurev@us.ibm.com}}

\maketitle

\begin{abstract}
Predictions for physical systems often rely upon knowledge acquired from ensembles of entities, e.g., ensembles of cells in biological sciences. For qualitative and quantitative analysis, these ensembles are simulated with parametric families of mechanistic models (MM). Two classes of methodologies, based on Bayesian inference and Population of Models, currently prevail in parameter estimation for physical systems. However, in Bayesian analysis, uninformative priors for MM parameters introduce undesirable bias. Here, we propose how to infer parameters within the framework of stochastic inverse problems (SIP), also termed data-consistent inversion, wherein the prior targets only uncertainties that arise due to MM non-invertibility. To demonstrate, we introduce new methods to solve SIP based on rejection sampling, Markov chain Monte Carlo, and generative adversarial networks (GANs). In addition, to overcome limitations of SIP, we reformulate SIP based on constrained optimization and present a novel GAN to solve the constrained optimization problem.
\end{abstract}

\section{Introduction}
\label{introduction}

\noindent In the era of machine learning (ML), mechanistic modeling is indispensable for performing prediction with datasets of small sizes, which are typical in biology, medicine and other physical sciences. Mechanistic modeling complements the power of pure ML by introducing formal model-based explanations of the data and additional prior information derived from known physical mechanisms. The first stage of generating predictions with mechanistic models is fitting a model to reproduce known physical samples, solving an \emph{inverse problem}. One widely used form of inverse problem estimates parameters in sets of observations acquired from single individuals. Another form of inverse problems, that we discuss in this paper, is where each data sample represents a single observation of an individual within an ensemble, and the goal is to find a parametric family of mechanistic models (MMs) that recreates the distribution of observations from the entire ensemble.

The above inverse problem can be reduced to a \emph{stochastic} inverse problem (SIP) \citep{butler2018combining}. In SIPs with deterministic MMs, the goal is to find a distribution $\mathcal{Q}_{X}$ of inputs to a function $\bm{y} = M(\bm{x})$, which, after ``push-forward'' by the function $M$, produces a given target distribution $\mathcal{Q}_{Y}$, such as $\bm{y} = M(\bm{x})\sim \mathcal{Q}_{Y},\; \bm{x} \sim \mathcal{Q}_{X}$. Usually the function $M$ is not invertible, so multiple (i.e., an infinite number of) input distributions can produce the target. Therefore, the prior $\mathcal{P}_{X}$ is introduced to restrict the solution using assumptions grounded in existing knowledge of parameter ranges and distributions.

A set of methods for $\mathcal{Q}_{X}$ estimation have been developed \citep{Poole2000}, termed ``consistent Bayesian inference'' \citep{butler2018combining}, or ``data-consistent inversion'' \citep{butler2020data-consistent}, which were initially developed for deterministic models \citep{butler2018combining} and have recently been extended to problems involving stochastic models  $\bm{y} = M(\bm{x}, \bm{\epsilon})$ \citep{butler2020data-consistent} with an additional vector of random variables, $\bm{\epsilon}$, modeling physical noise. These methods are based on Monte Carlo sampling, importance sampling, rejection sampling, and density estimation. However, several major drawbacks prevent such methods from being widely applied in SIPs, including diminishing accuracy as dimensionality increases and difficulty handling the type of complex inference scenarios and arbitrary data distributions that present themselves in practice. In the current work, we aim to overcome these challenges by reformulating SIP as an optimization problem. We present new methods to solve SIP by adapting previous rejection sampling approaches, as well as introducing deep learning into SIP for the first time to solve our optimization problem, opening new avenues for solving SIPs.

The first novel algorithm we introduce is a sequential rejection algorithm that performs data-consistent inversion \citep{butler2020data-consistent} by employing a series of optimization steps. We refine this method using Markov chain Monte Carlo (MCMC) to initialize a proposal distribution prior to rejection. These approaches, described in Methods, form new benchmarks for SIP using conventional ML methods, which we can use for comparison and evaluation of our novel deep learning networks.

Although deep learning has not previously been used in SIP, multiple deep learning methods have been recently developed for application to the related problem of model inversion for single observations, termed the parameter identification problem (PIP) or simulation-based inference (SBI) \citep{cranmer2020frontier, goncalves2020training, lueckmann2021benchmarking}. These include conditional networks such as conditional normalizing flows \cite{Papamakarios2017, papamakarios2019sequential, papamakarios2021normalizing} and the conditional generative adversarial network (c-GAN) \citep{adler2018deep, ramesh2022gatsbi}. Therefore, we first tested the c-GAN\cite{mirza2014conditional} to estimate $\mathcal{Q}_{X}$, while identifying some of the limitations that this approach shares with prior SIP methods. We then show that more complex SIPs and SIPs with stochastic MMs can be solved when formulated as constrained optimization problems using a novel generative model architecture, described in Methods, which we implement here using GANs. 

We experimentally evaluate our methods on a range of example problems and demonstrate equivalent performance to the benchmark method of rejection sampling. Next, we tackle problems that are not accessible to existing methods and therefore analyze these deep learning results without comparison. For an intuitive example of a high-dimensional parameter inference problem involving highly structured data, we present super-resolution imaging on the MNIST dataset reformulated as a SIP. In Supplementary Information, we demonstrate the ability to extend our GAN architecture to solve a parameter inference problem in an intervention scenario, analogous to drugging a population of cells in a biological experiment. Our results establish GANs as a new instrument that can solve configurations of parameter inference problems for which there are currently no solutions.

\section{Methods}
\label{methods}

\subsection{Background and Related Work}
\label{background}

In the inverse problem, observable vector-valued functions $t\mapsto s_{\tau}(t)$, e.g., sets of experimental signal waveforms $\{s_{\tau}:\tau \in J\}\subseteq S$, are recorded from objects in an ensemble. The functions of time $s_{\tau}(t)$ are indexed by a discrete or continuous index $\tau\in J$. We are interested in finding solutions $\{f(t;\bm{x},\bm{\epsilon}): \bm{x}\in\mathbb{R}^n, \bm{\epsilon}\in\mathbb{R}^d\}\subseteq S$ of differential equations for a given MM that approximate the experimental observations. Here $\bm{x}$ is a vector of MM parameters, $\bm{\epsilon}$ is a vector of random variables and $S$ is a functional space of continuous time signals. Feature vectors $L(s_{\tau}(\cdot))$ and $L(f(\mathord{\cdot};\bm{x},\bm{\epsilon}))$ (also referred to as quantities of interest) are extracted from experimental $s_\tau$ and simulated $f(\cdot;\bm{x},\bm{\epsilon})$ signals using some map $L:S\to \mathbb{R}^m$. In the SIP, the model is usually reduced to a function $\bm{y}=M(\bm{x},\bm{\epsilon})$, defined as $M(\bm{x},\bm{\epsilon})=L(f(\mathord{\cdot};\bm{x},\bm{\epsilon}))$. If the objects are approximated by deterministic MMs, then $\bm{y}=M(\bm{x})$. The goal is to find the distribution of MM parameters $\mathcal{Q}_X$, which when passed through $M$, generates a distribution of outputs matching the distribution of features $\mathcal{Q}_Y$ extracted from experimental signals $L(s_{\tau}(\cdot))$. The model function $M$ could be in a closed form or represented by a surrogate trained on features from numerical solutions of MM differential equations. Therefore in analysis, we use the function $M$ rather than the original differential equation model and refer to $M$ as the ``model''. Feature extraction functions $L$ are typically hand-crafted, but recent work has proposed automated feature extraction within the SIP context \citep{mattis2022learningQOI}. It should be emphasized that SIP is different from parameter identification problems that aim to estimate parameters for observed data from a single individual, which are often solved using classical Bayesian inference. These two different parameter inference objectives have recently been compared in detail \citep{pilosov2023parameterestimation}. In general, when sampling from the posterior using Bayesian inference, samples will be biased by the prior and, when passed through the model, will not generate samples from the target distribution (for empirical comparison, see \citep{butler2018combining,pilosov2023parameterestimation}).

In the case of deterministic MMs, Poole and Raftery \cite{Poole2000} showed that given random variables $X$ and $Y$, linked deterministically by $\bm{y}=M(\bm{x})$, the density $q_Y(\bm{y})$ of function outputs can be mapped to the density of function inputs $q_X(\bm{x})$ coherent to the input data using the equation
 \begin{equation} \label{eq:poole_raftery}
     q_X(\bm{x}) \equiv q_Y(\bm{y})\rightbar*{\frac{p_X(\bm{x})}{p_Y(\bm{y})}}_{\bm{y} = M(\bm{x})},
 \end{equation}
where $p_X(\bm{x})$ is the prior density on the input, and $p_Y(\bm{y})$ is the model-induced prior density obtained upon sampling from $p_X(\bm{x})$ and applying the function $M$ to the samples (push-forward of the prior). For invertible functions, the ratio $p_X(\bm{x})\mathbin{/}{p_Y(M(\bm{x}))}$ is simply the Jacobian of the function $M$. Recent work \cite{butler2018combining}, which rediscovered Poole and Raftery's formulation, outlines direct parallels between \eqref{eq:poole_raftery} and classical Bayesian inference and provides measure theoretic proofs of important properties of \eqref{eq:poole_raftery}, such as solution stability. Modifications \citep{butler2020data-consistent} of \eqref{eq:poole_raftery} (termed `data-consistent inversion') generalized this framework to stochastic models and extended its applicability to a wider variety of SIPs with stochastic $\bm{y} = M(\bm{x},\bm{\epsilon)}$,
\begin{equation} \label{eq:poole_raftery_stoch}
     q_{X,E}(\bm{x}, \bm{\epsilon}) \equiv q_Y(\bm{y})\rightbar*{\frac{p_{X,E}(\bm{x}, \bm{\epsilon})}{p_Y(\bm{y})}}_{\bm{y} = M(\bm{x},\bm{\epsilon})}.
 \end{equation}
After sampling from $q_{X,E}$, the marginal samples are taken by dropping $\epsilon$. Note that there is no difference in treatment of model parameters $\bm{x}$ and noise $\bm{\epsilon}$ in \eqref{eq:poole_raftery_stoch}. In real-world problems, the nature of noise is often assumed a priori (e.g., Gaussian), and noise is assumed to be independent of mechanistic model parameters. However, $\bm{x}$ and $\bm{\epsilon}$ can become correlated in samples from $q_X(\bm{x}, \bm{\epsilon})$ \citep{butler2020data-consistent}. In this paper, we present a method for configuring a parametric model of the noise during estimation of a data-consistent solution $q_X(\bm{x})$ to the general SIP with stochastic models.

Two methods are proposed in \citep{butler2018combining, butler2020data-consistent} to sample from $q_X$. First, the density $p_Y(\bm{y})$ can be estimated by some standard density estimator and parameters can be sampled by Monte Carlo methods from \eqref{eq:poole_raftery} and \eqref{eq:poole_raftery_stoch}. The second option \citep{butler2018combining} is a rejection sampling algorithm (algorithm \ref{alg:rejection_orig}). For conciseness, we provide the version of the algorithm for deterministic models, since the methods to solve \eqref{eq:poole_raftery} and \eqref{eq:poole_raftery_stoch} are essentially the same. 
\begin{algorithm}
    \caption{Rejection sampling \citep{butler2018combining}}
    \label{alg:rejection_orig}
    \begin{algorithmic}[1]
        \Require Number of samples and iterations $N_S$; model $\bm{y}=M(\bm{x})$; density  $ q_{Y} $
            \State Sample $\left\{\bm{x}_i\right\}_{i=1}^{N_S} \sim p_{\bm{X}} $
            \For{i=1...$N_S$}
                \State $ \bm{y}_i := M(\bm{x}_i)$
            \EndFor
            \State $P:= \left\{\bm{x}_i, \bm{y}_i\right\}_{i=1}^{N_S}$
            \State Use $\{\bm{y}_i : (\bm{x}_{i}, \bm{y}_{i}) \in P\}$ to build density model $\hat{p}_Y$
            \State Compute $B: = \max_i \frac{q_Y(\bm{y}_i)}{\hat{p}_Y(\bm{y}_i)}$
            \For{$(\bm{x}_i, \bm{y}_i) \in P$}
                \State $ \rho \sim \mathcal{U}\left([0, 1]\right) $
                \If{$\rho > \frac{q_Y(\bm{y}_i)}{B\times \hat{p}_Y(\bm{y}_i)}$}
                \State Reject $P: =P \setminus (\bm{x}_{i}, \bm{y}_{i})$
                \EndIf
            \EndFor
            \State \Return {$\big\{\bm{x}_{i}: (\bm{x}_{i}, \bm{y}_{i}) \in P\big\}$}
    \end{algorithmic}
\end{algorithm}
Another algorithm based on importance sampling was proposed in \citep{Poole2000}. The main limitation of all above methods is high instability and variance due to inaccuracy in estimation of $p_Y(\bm{y})$, since this term appears in the denominators. For example, during rejection sampling, errors in estimation may lead to a large, possibly infinite, bound constant $B$ (step 7 in algorithm \ref{alg:rejection_orig}) and cause a high rejection rate.

In parallel with the methods developed in \citep{butler2018combining} following \citep{Poole2000}, an empirically-derived method to solve a similar problem was described in \citep{lawson2018unlocking}. There, a sequential Monte Carlo stage was used to sample from $\left. q_Y(\bm{y})p_{X}(\bm{x})\right\vert_{y=M(x)}$ with uniform $p_{X}(\bm{x})$, obtaining initial samples that are then refined using a rejection stage that reduces JS-divergence (JSD) between the samples and the target observations. This solution did not rely on the theoretical background of \citep{butler2018combining}, but solves a similar problem using different methods.

Deep learning methods have not been applied for parameter inference in SIP. However, deep learning is widespread in the PIP/SBI problems mentioned above, wherein experimental data $\mathcal{D} = \{\bm{y}_i\}$ is acquired from a single individual over several trials, and simulated data derives from a parameterizable, stochastic model with parameters $\bm{x}$ and outputs $\bm{y}$ simulating $p(\bm{y}|\bm{x})$. SBI is commonly solved using Approximate Bayesian Computation (ABC), or ``likelihood-free'' methods, to bypass the intractable likelihood term \citep{cranmer2020frontier}. The task is to infer the posterior distribution $\mathcal{P}(\bm{x}|\mathcal{D})$ of model parameters $\bm{x}$ that likely generated the individual's observed data. This scenario can be formulated for independent observations as:
\begin{equation}
        p(\bm{x}|\mathcal{D}) =  \frac{p(\mathcal{D}|\bm{x})  \cdot p(\bm{x})}{p(\mathcal{D})} \propto \prod_i p(y_i|\bm{x})  \cdot p(\bm{x}).
\label{eq:Bayesinf}
\end{equation}
Recent advances in conditional neural density estimation have enabled a host of new approaches in the SBI domain to approximate the likelihood and posterior densities with networks trained on simulated data. Mixture density networks \citep{papamakarios2016fast,goncalves2020training}, emulator networks \citep{lueckmann2019likelihood, Fengler2020} and autoregressive flows \citep{papamakarios2019sequential,goncalves2020training} each approximate either the likelihood $p(\bm{y}|\bm{x})$ or posterior $p(\bm{x}|\bm{y})$. Models of the posterior $p(\bm{x}|\bm{y})$ are trained on samples $(\bm{x}_i, \bm{y}_i)$ by sampling $\bm{x}_i \sim \mathcal{P}_{X}$ and running the MM $\bm{y_i} \sim p(\bm{y}| \bm{x}_i)$. The most advanced networks for parameter inference from the literature are conditional normalizing flows \cite{radev2020bayesflow}, which perform invertible transformations of random variables. Invertibility requires the determinant Jacobian to be cheaply computable, which limits the range of transformations that can be used. Advances such as masked autoregressive flows (MAF) \citep{Papamakarios2017}, inverse autoregressive flows (IAF) \citep{kingma1606improving} and neural autoregressive flows (NAFs) \citep{huang2020augmented} have each improved the accuracy of inference achievable with normalizing flow models. These conditional density estimation networks provide explicit density models that can be used for fast approximation of density from multiple datasets $\mathcal{D}$ in SBI problems. Following the same trend, we tested the c-GAN as an implicit estimator of stochastic maps for MM parameter inference in SIP. Other conditional distribution models, such as conditional diffusion models, could be used instead of c-GAN for this purpose.

Although conditional neural networks are suitable for parameter inference, they are essentially estimators of distributions from previously sampled sets of parameters, in contrast to other methods that actively explore parameter space and propose parameter samples for evaluation. These networks may require a large sample size for training and capacity to learn the conditional distribution, and inference accuracy may suffer in regions of $\bm{y}$ that have low density in $\mathcal{P}_{Y}$ (the model induced prior), necessitating sequential approaches \citep{papamakarios2019sequential}. In practice, however, conditional networks often show good performance even in high-dimensional parameter inference scenarios \citep{ramesh2022gatsbi}. For SIP with stochastic MMs c-GAN provides the same solution as Rejection methods applied to \eqref{eq:poole_raftery_stoch}. Here, we propose a reformulation of SIP as an optimization problem to facilitate the use of deep neural networks as a flexible approach to a broader class of SIP applications (see Results section B).
\begin{algorithm}[] 
\caption{Iterative rejection sampling}
\label{alg:rejection}
    \begin{algorithmic}[1] 
    \Require Number of samples and iterations $N_S$, $N_I$; constant $B$, model $\bm{y}=M(\bm{x})$; density  $ q_{Y} $
        \State Sample $ \left\{\bm{x}_i\right\}_{i=1}^{N_S} \sim p_{\bm{X}} $
        \For{i=1...$N_S$}
            \State $ \bm{y}_i := M(\bm{x}_i)$
        \EndFor
        \State $P:=\left\{\bm{x}_{i}, \bm{y}_{i}\right\}_{i=1}^{N_S} $
        \For{j=1...$N_I$}
            \State Set $Q:= \big\{\bm{y}_{i}: (\bm{x}_{i}, \bm{y}_{i}) \in P\big\}$
            \State Use $Q$ to build density estimator $\hat{q}_Y$, e.g., using GMM
            \For{$(\bm{x}_{i}, \bm{y}_{i}) \in P$} 
                \State $ \lambda_{i} = \min\left( \frac{q_{Y}(\bm{y}_{i})}{B \times \hat{q}_Y(\bm{y}_{i})}, 1 \right) $
                \State $ \rho \sim \mathcal{U}\left([0, 1]\right) $
                \If{$\rho > \lambda_{i}$}
                    \State Reject $P: =P \setminus (\bm{x}_{i}, \bm{y}_{i})$
                \EndIf
            \EndFor
        \EndFor
    \State \Return{$\big\{\bm{x}_{i}: (\bm{x}_{i}, \bm{y}_{i}) \in P\big\}$} 
    \end{algorithmic} 

\end{algorithm}

\subsection{Parameter inference methods for SIP}
\label{ParamInfGAN}

We present four novel methods to solve SIP. We use the first two, a modification of the rejection sampling (algorithm \ref{alg:rejection_orig}), and rejection sampling initialized with an MCMC stage, as numerical benchmarks for comparing the results of the subsequent GAN-based solutions.

\subsubsection{Modified rejection sampling}

A substantial problem with algorithm \ref{alg:rejection_orig} is the possibility of a large bound constant $B$. To address this, we modified the algorithm to perform rejections in multiple iterations, selecting samples for rejection to incrementally minimize divergence between model outputs and observations at each iteration, as described in algorithm \ref{alg:rejection}. The main difference from algorithm \ref{alg:rejection_orig} is that the constant $B$ does not have to be a bounding constant of the density ratio, but a parameter of the algorithm. If $B$ is equal to the bounding constant, then $N_I=1$ and algorithm \ref{alg:rejection} reduces to algorithm \ref{alg:rejection_orig}. We used $B=1$ and $N_I = 10$ in all our examples, which was sufficient for $\hat{q}_Y(\bm{y})$ to converge to $q_Y(\bm{y})$, resulting in distributions consistent with examples in \citep{butler2018combining,butler2020data-consistent}. In real-world applications, algorithm \eqref{alg:rejection} can be modified by defining a target number of samples to reject at each iteration, with $B$ calculated to reject the expected number of samples. If the algorithm is not converged, $N_I$ can be increased and additional points sampled from the prior to increase initial sample size. However, we have not explored such strategies here and use fixed values of $B$ and $N_I$ across all examples. We use algorithm \ref{alg:rejection} as our first baseline method.

\subsubsection{Rejection sampling boosted with MCMC}

In algorithms \ref{alg:rejection_orig} and \ref{alg:rejection}, the prior $p_{X}$ is used to generate proposal sample sets. As discussed in \citep{Poole2000}, there is an equivalence class of priors that induce identical target density. We can therefore consider an arbitrary proposal distribution with density $p_p(\bm{x})$ such that for any pair $(\bm{x_1}, \bm{x_2}) \in \{(\bm{x_1}, \bm{x_2}): M(\bm{x_1})=M(\bm{x_2})\}$, 
\begin{equation*}
    \frac{p_p(\bm{x_1})}{p_p(\bm{x_2})} = \frac{p_{X}(\bm{x_1})}{p_{X}(\bm{x_2})} ,
\end{equation*}
and with support of the push-forward of $p_p(\bm{x})$ that covers the support of $p_Y(\bm{y})$. Such a proposal distribution could be any distribution with a PDF proportional to the function $\left. g(\bm{y})p_{X}(\bm{x})\right\vert_{y=M(x)}$, where $g(\bm{y})$ is an arbitrary function, $g(\bm{y}) \geq 0$ with strict inequality in $\bm{y}$ where $p_Y(\bm{y}) > 0$. Examples of such proposal distributions could be simply the prior $p_{X}(\bm{x})$, or the numerator of \eqref{eq:poole_raftery}, $\left. q_Y(\bm{y})p_{X}(\bm{x})\right\vert_{y=M(x)}$. The latter is often closer to the solution, and we used it, sampled with MCMC, to boost the rejection method by providing a higher density of samples in the regions of interest from which to construct the initial density model in algorithm \ref{alg:rejection}. 

\subsubsection{Conditional GAN for amortized inference}

For deterministic models $\bm{y}=M(\bm{x})$, we are able to use the c-GAN \cite{mirza2014conditional} to find MM parameter distributions given experimental observations. In this case, $p(x |y)$ is ill-defined, since the distribution of parameters that correspond to $\bm{y}$ is degenerate. As \cite{goodfellow2014generative} noted, one advantage of GANs is their ability to represent degenerate distributions, and we were able to apply c-GAN in SIP with deterministic models. The generator network of c-GAN, $\bm{x} = G(\bm{y}, \bm{z};\theta)$, with network parameters $\theta$, which generates $\bm{x}$ from $\bm{y}$ and samples $\bm{z}\sim \mathcal{P}_Z$ of the base GAN distribution, is trained on the samples $(\bm{x}_i, \bm{y}_i)$, where $\bm{x}_i \sim p_X(\bm{x})$ and $\bm{y}_i=M(\bm{x}_i)$. At the inference stage, $\bm{y}_i \sim \mathcal{Q}_{Y}$ would be transferred to samples from estimated $\mathcal{Q}_{X}$ by $\bm{z}_i\sim \mathcal{P}_Z$, $\bm{x}_i = G(\bm{y_i}, \bm{z}_i)$. The advantage of this approach is that it uses `amortized inference' \cite{cranmer2020frontier}, providing solutions to any number of problems at the inference stage after being trained a single time using one simulated training data set. In several examples, we demonstrate that results obtained using c-GAN are empirically consistent with results obtained from \eqref{eq:poole_raftery} and \eqref{eq:poole_raftery_stoch} using algorithm \ref{alg:rejection}. 

\subsubsection{Regularized generative adversarial networks}

\begin{figure*}
  \centering
  \includegraphics[width=.8\linewidth]{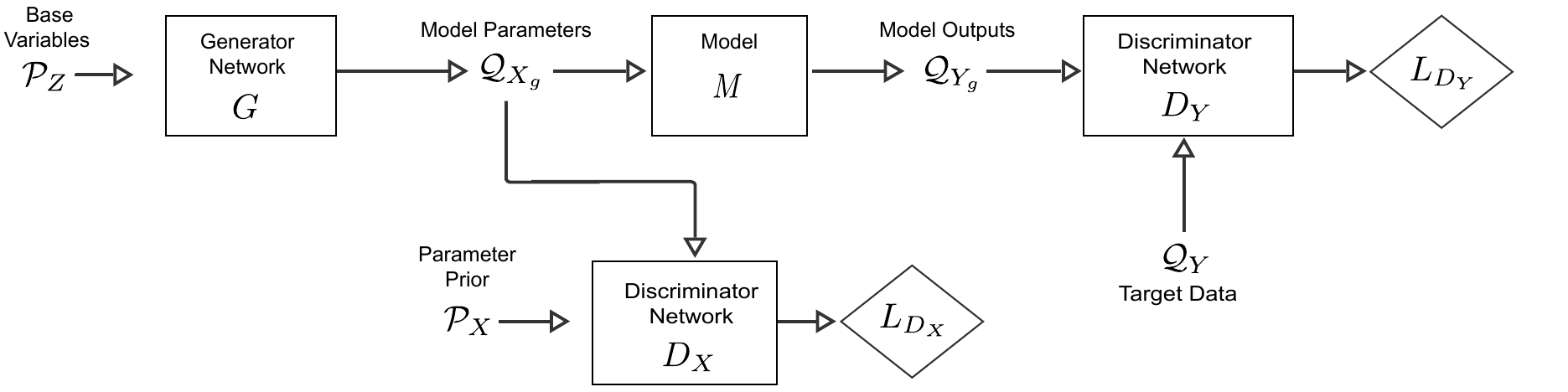}
  \caption{Generative adversarial models for inference of model input parameters. Regularized GAN (r-GAN) solves the constrained optimization problem using the penalty method. The loss of the generator is the weighted sum of losses from r-GAN's two discriminators. r-GAN enforces the equality of $\mathcal{Q}_Y$ and $\mathcal{Q}_{Y_g}$ and maximizes the overlap between $\mathcal{P}_X$ and $\mathcal{Q}_{X_g}$.}
  \label{fig:rgan}
\end{figure*}

In \eqref{eq:poole_raftery}, the prior density $p_X(\bm{x})$, as in Bayes formula, is used as the relative likelihood of model input parameter values. Here, we reformulate the problem as a constrained optimization, aiming to minimize divergence between the prior $\mathcal{P}_X$ and the distribution $\mathcal{Q}_{X_g}$ of parameters sampled by a generator in a GAN from some parametric family $G_\theta\in\{G_\theta(\cdot) | \theta \in \Theta\}$, given that the distribution $\mathcal{Q}_{Y_g}$ of the push-forward of $\mathcal{Q}_{X_g}$ through the model $M$ matches the target distribution of observations $\mathcal{Q}_Y$ (the primary goal in SIPs). This places a regularizing constraint on parameters sampled by the generator, in a model we term the regularized generative adversarial network (r-GAN) solution to SIP, shown in Figure \ref{fig:rgan}. Thus, the problem is formulated as
\begin{equation} \label{eq:coGAN}
\begin{aligned}
& \text{given}\; & & \mathcal{P}_{X},\; \mathcal{Q}_{Y},\; \bm{y}=M(\bm{x}) \\ 
& \text{minimize}
& & D_f(\mathcal{Q}_{X_g} || \mathcal{P}_X) \\
& \text{subject to}
& & supp(X_g)\subseteq supp(X),\;D_f(\mathcal{Q}_{Y_g} || \mathcal{Q}_Y) = 0 \\
& \text{where}\; & &  \bm{y}_g = M(\bm{x}_g)\sim \mathcal{Q}_{Y_g},\; \bm{x}_g \sim \mathcal{Q}_{X_g}. \\ 
\end{aligned}
\end{equation}
In \eqref{eq:coGAN}, $D_f(\cdot||\cdot)$ is an f-divergence measure such as Jensen-Shannon (JS) divergence. This reformulation of the problem provides another way to account for the prior. We are looking for not just any distribution of model input parameters that produces $\mathcal{Q}_Y$, but the distribution with minimal divergence from the prior. The additional constraint $supp(X_g)\subseteq supp(X)$ ensures that the distribution of the generated input parameters $X_g$ is within the prior bounds. To solve \eqref{eq:coGAN}, we first convert the constrained optimization problem to a non-constrained problem with minimization of $w_{X}\times D_f(\mathcal{P}_X || \mathcal{Q}_{X_g}) + w_{Y}\times D_f(\mathcal{Q}_{Y_g} || \mathcal{Q}_Y)$, solving \eqref{eq:coGAN} with the penalty method. In a series of training steps, the weights  $w_{Y}$ and $w_{X}$ are changed such that the weight $w_{X}$ is much smaller than the weight $w_{Y}$ at the last step. For example, we used $w_{Y} = 0$, $w_{X} = 1$ and $w_{Y} = 1.0$, $w_{X} = 0.1$ for the first and last steps, respectively. Non-constrained optimization could be solved with different methods including methods that resemble sequential Monte Carlo sampling \citep{DelMoral2021smc}, perturbing and resampling particles that represent MM parameter distributions in a series of importance sampling iterations. Rather than optimization by stochastically perturbing samples, we instead sample using a neural network generator, minimizing the divergence $D_f(\mathcal{P}_X || \mathcal{Q}_{X_g})$ over $\theta$ in the generator: $\bm{z}\sim \mathcal{P}_Z$, $\bm{x}_g = G_\theta(\bm{z})\sim \mathcal{Q}_{X_g}$, where $\mathcal{P}_Z$ is a base distribution, usually Gaussian. We can then optimize with stochastic gradient descent, and employ GAN discriminators to calculate the divergence measures in the optimization problem. The r-GAN has two discriminators, and the generator loss $L_{G}$ is composed of a weighted sum of losses due to both discriminators, generator loss $L_{D_{X}}$ due to discriminator $D_X$ and loss $L_{D_{X}}$ due to $D_Y$,
\begin{equation}
    \begin{split}
        L_{G} = w_{Y}L_{D_{Y}} + w_{X}L_{D_{X}}.
    \end{split}
\label{eq:LG}
\end{equation}
Different f-divergence measures could be applied using different GAN loss functions \cite{nowozin2016f}. This type of network is similar to that used in adversarial variational optimization (AVO) \citep{louppe2019avo}, where adversarial optimization minimizes $D_f(\mathcal{Q}_{Y_g} || \mathcal{Q}_Y)$. However, a major difference is the use of the second discriminator to place a regularization on the generated samples, whereas AVO uses a penalty term that instead decreases sample entropy. Another discriminator-based method for model inversion has been proposed \citep{jethava2018ABCGAN} in the context of ABC, but no regularization according to the parameter distribution is used in this example, and SIP solutions are not addressed.

Our SIP reformulation using r-GAN provides several benefits over existing methods. r-GAN makes more accurate predictions for general stochastic SIPs, demonstrated by our stochastic MM experiment, enables sampling from distributions of high-dimensional data as demonstrated by our imaging experiment, and can be extended to complex simulation configurations as demonstrated in our intervention experiment and additional experiments in Supplementary Information. 

\section{Results}

To illustrate our method's applicability to a variety of mechanistic models, we used four test functions as the mechanistic model $\bm{y} = M(\bm{x})$ and solved SIP for each using rejection sampling, MCMC-boosted rejection sampling, c-GAN, and r-GAN methods. The first example was used as a standard example of SIP in \cite{butler2018combining} and \cite {breidt2011measure}, and we use it here to compare our SIP methods and highlight the role of the prior discriminator. For the remaining three examples, r-GAN is the only existing solution, and we therefore demonstrate scenarios that could not previously be solved accurately. Second, we use SIP with a stochastic simulator, which was attempted in a previous publication using algorithm \ref{alg:rejection_orig} \cite{butler2020data-consistent}, and for which we show that r-GAN improves upon biased solutions produced using the other methods. Third, we use r-GAN for a super-resolution imaging SIP, an example of a high-dimensional implicit sampling problem with no analogous solution among other methods. Fourth, in the Supplementary Information, we demonstrate SIP for an intervention scenario using an extension to r-GAN, which highlights the potential usefulness of this extensible framework for tackling complex SIP configurations. To quantify performance, we estimated both $\text{JSD}(\mathcal{P}_{X}||\mathcal{Q}_{X_g})$ and $\text{JSD}(\mathcal{Q}_{Y}||\mathcal{Q}_{Y_g})$ using classifiers trained on samples from the distributions. We provide summary implementation details here and full details on the GAN implementation, GAN configuration, JS-divergence estimation from implicit samples and MCMC methods in the Supplementary Information. Demonstration code for r-GAN in Pytorch is available at https://github.com/IBM/rgan-demo-pytorch. For additional deterministic function tests from the literature, see Supplementary Information.

For a fair comparison between inference methods, we used a fixed computational budget of 2,000,000 simulations of the mechanistic model. For Rejection and c-GAN, these simulations were performed in advance, generating samples used to initialize the algorithms. In MCMC-boosted rejection, we perform simulations throughout iterations of MCMC, (algorithm specifics are provided in the Supplementary Information). The r-GAN requires simulation of the mechanistic model during training for each sample from the generator network, and we trained the r-GAN for $200$ epochs with $10,000$ samples per epoch. We did not perform elaborative computational benchmarks between methods, as we envision the methods being used in combination according to their strengths and weaknesses for specific problems.

\subsection{Non-linear function and the role of the prior}

\begin{figure*}[htbp!]
  \centering
  \includegraphics[width=0.8\textwidth]{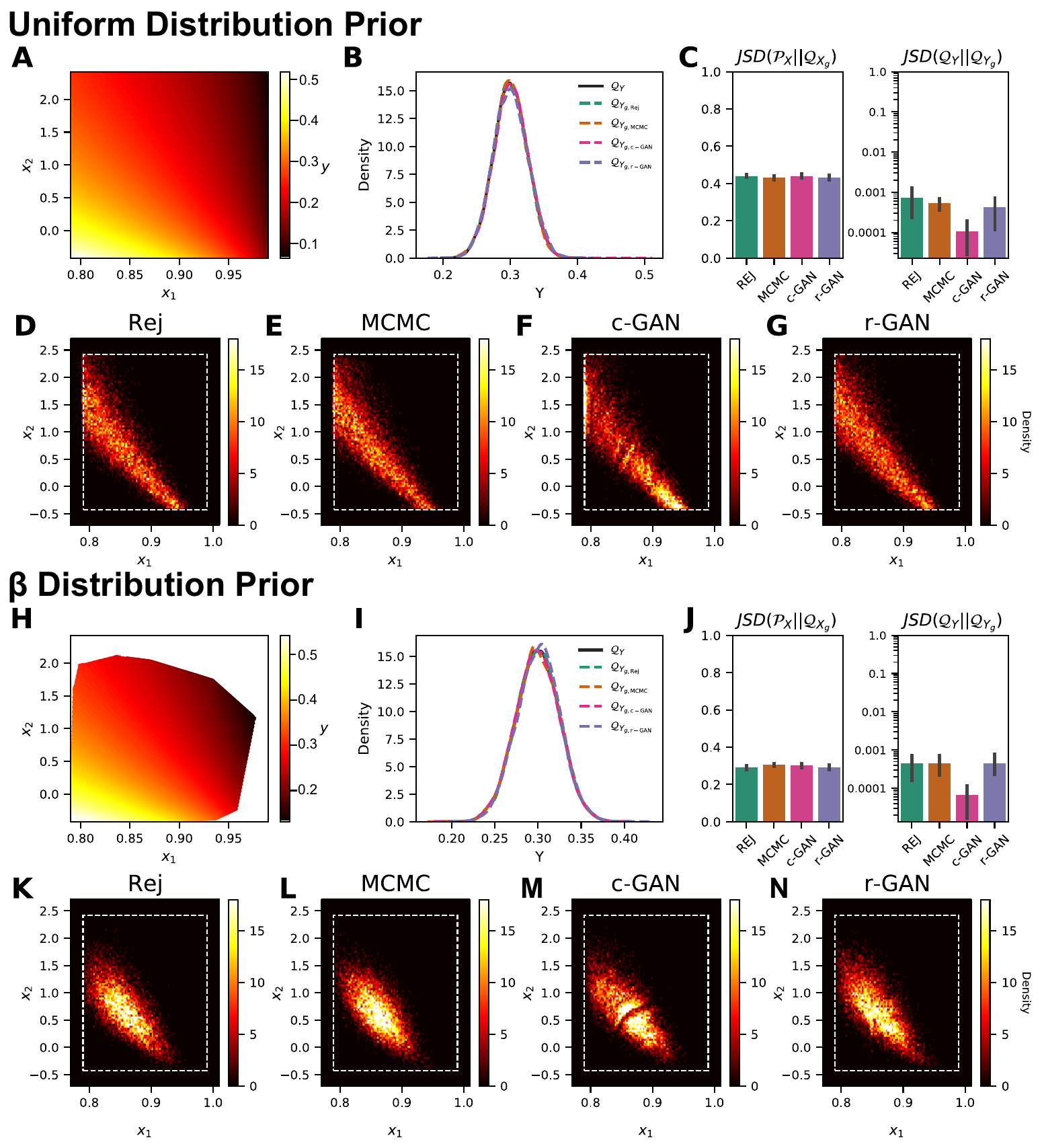}
  \caption{Parameter inference for \eqref{eq:system} with two different prior distributions. \textbf{A}. Heat map of $y_2$ over uniform $\mathcal{P}_X$. \textbf{B}. KDE of $\mathcal{Q}_Y$ and the generated (inferred) output distributions $\mathcal{Q}_{Y_g}$ using Rejection, MCMC, c-GAN and r-GAN. \textbf{C}. Left: estimated JS-divergence between $\mathcal{P}_X$ and $\mathcal{Q}_X$. Right: estimated JS-divergence between $\mathcal{Q}_Y$ and $\mathcal{Q}_{Y_g}$. \textbf{D-G} Histograms of $\mathcal{Q}_{X_g}$ for each method. \textbf{H-N} as in \textbf{A-F}, but for $\beta$ distribution prior. In \textbf{D-G} and \textbf{K-N}, dashed rectangle denotes bounds set by $\mathcal{P}_X$.}
  \label{fig:nonlinear1}
\end{figure*}

In the first example, we represented the mechanistic model by a nonlinear system of equations with two input parameters ($x_1$ and $x_2$), 
\begin{equation} \label{eq:system}
\begin{aligned}
  & x_1 y_1^2 + y_2^2 = 1 \\
  & y_1^2 - x_2 y_2^2 = 1,
\end{aligned}
\end{equation}
with $y_2$ as the model output (shown in Figure \ref{fig:nonlinear1}A as $\bm{y}$). The target observation distribution $\mathcal{Q}_Y$ was synthetic data with distribution $\mathcal{N}(0.3, 0.025^2)$, truncated to the interval $(0, 0.6)$ (Figure \ref{fig:nonlinear1}B, black). 

We used this problem to assess the influence of the parameter prior on generated samples using two different $\mathcal{P}_X$. First, a uniform prior was considered with $x_1 \sim \pazocal{U}(0.79, 0.99)$ and $x_2 \sim \pazocal{U}(1 - 4.5\sqrt{0.1}, 1 + 4.5\sqrt{0.1})$ as in \cite{butler2018combining}. Figure \ref{fig:nonlinear1}B shows the distribution of $y_2$ (after push-forward with \eqref{eq:system}) obtained using Rejection, MCMC, c-GAN, and r-GAN to sample $x_1$ and $x_2$, with histograms of parameter samples for each method shown in Figures \ref{fig:nonlinear1}D, E, F, and G, respectively. JSD between the prior $\mathcal{P}_X$ and the generated samples $\mathcal{Q}_X$, and between the push-forward of the generated samples through the model $\mathcal{Q}_{Y_g}$ and the target output distribution $\mathcal{Q}_Y$ (Figure \ref{fig:nonlinear1}C) indicated that all four methods performed well in estimating model input parameters coherent to the target distribution. While the c-GAN approach has lower $\text{JSD}(\mathcal{Q}_Y||\mathcal{Q}_{Y_g})$ (Figure \ref{fig:nonlinear1}), indicating slightly more accurate sampling in model output space, it also has more artificial structure in parameter samples (Figure \ref{fig:nonlinear1}F) than other approaches, likely due to the generator network learning the full conditional distribution $\mathcal{P}_X,Y$ required for amortized inference.

Next, we altered the prior over $x_1$ and $x_2$ to $x_1 \sim \text{Beta}(2, 5, 0.79, 0.99)$ and $x_2 \sim \text{Beta}(2, 5, 1 - 4.5\sqrt{0.1}, 1 + 4.5\sqrt{0.1})$,  as in \cite{butler2018combining}. Figure \ref{fig:nonlinear1}G shows $\bm{y}$ with limits imposed by samples from this new prior distribution. The joint distributions of $x_1$ and $x_2$ (Figures \ref{fig:nonlinear1}K-N) demonstrate that the inferred parameter distribution is influenced by $\mathcal{P}_X$ (when compared with Figures \ref{fig:nonlinear1}D-G), while, after push-forward through the model to $\mathcal{Q}_{Y_g}$, the inferred input parameter samples are coherent with the target observations $\mathcal{Q}_Y$ for both distinct priors (Figures \ref{fig:nonlinear1}B and \ref{fig:nonlinear1}I).

\subsection{SIP with a stochastic model}

\begin{figure*}[htbp!]
  \centering
  \includegraphics[width=0.8\linewidth]{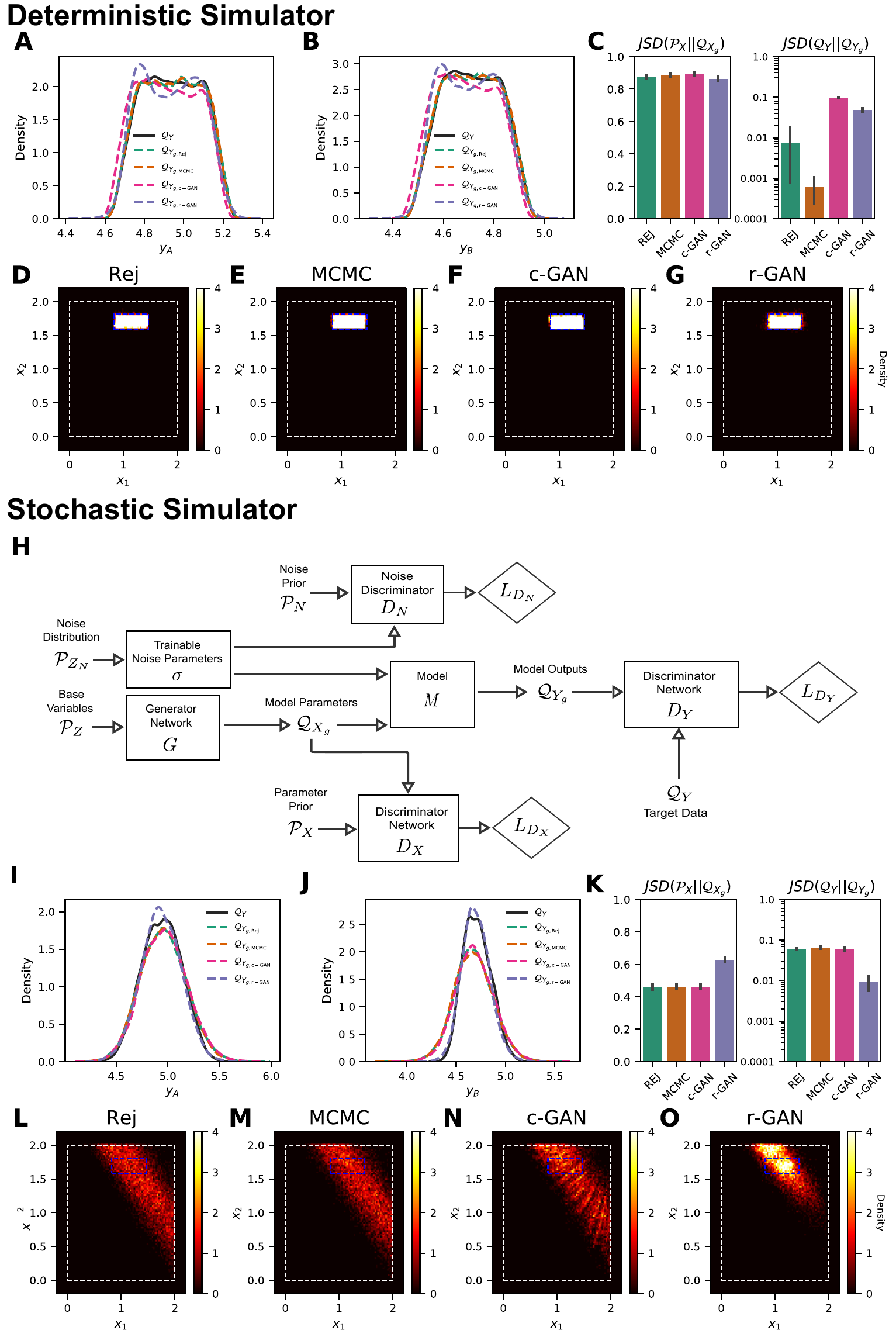}
  \caption{Parameter inference for stochastic simulator \eqref{eq:wobblyplate} \textbf{A-B}. Marginal KDEs of ($y_A$, $y_B$) constituting $\mathcal{Q}_Y$ and the generated (inferred) output distributions $\mathcal{Q}_{Y_g}$ using Rejection, MCMC, c-GAN and r-GAN. \textbf{C}. Left: estimated JS-divergence between $\mathcal{P}_X$ and $\mathcal{Q}_X$. Right: estimated JS-divergence between $\mathcal{Q}_Y$ and $\mathcal{Q}_{Y_g}$. \textbf{D-G} Histograms of $\mathcal{Q}_{X_g}$ for each method. \textbf{H}. Schematic diagram of r-GAN for stochastic MMs. Note the additional generator and discriminator for noise parameters compared with the r-GAN in Fig. \ref{fig:rgan}. \textbf{I-O}. As in \textbf{A-G}, but for stochastic model, with embedded noise using $\epsilon_1$ and $\epsilon_2$ in \eqref{eq:wobblyplate}. In \textbf{D-G} and \textbf{L-O}, white dashed rectangle denotes bounds set by $\mathcal{P}_X$, and blue dashed rectangle denotes bounds of uniform `true' parameter distribution.}
  \label{fig:wobblyplate}
\end{figure*}

To test SIP using GANs with a stochastic mechanistic model, we used a model of a tiltmeter, as in \cite{butler2020data-consistent}. Here, a square plate is wobbling around a pivot positioned near the origin, and the height of the plate $y$ at a location on the surface ($p_1$, $p_2$) is given by
\begin{equation} \label{eq:wobblyplate}
\begin{aligned}
  y = y_0 + x_1 (p_1 + \epsilon_1) + x_2 (p_2 + \epsilon_2),
\end{aligned}
\end{equation}
where $y_0 = 3.0$ is the height of the plate above the origin, ($x_1$, $x_2$) are the slopes of the plate and the parameters to be inferred, and ($\epsilon_1$, $\epsilon_2$) are stochastic noise embedded in the simulator, effectively perturbing the measurement location. In a deterministic scenario, the height of the plate is measured with complete accuracy (($\epsilon_1$, $\epsilon_2$) = ($0$, $0$)) at two positions $\bm{P_A} = (0.6, 0.6)$ and $\bm{P_B} = (0.8, 0.6)$ at different times to form the target $\mathcal{Q}_Y$, comprising measurements ($y_A$, $y_B$). The parameter prior was considered with $x \sim \pazocal{U}(0, 2)$ for $x_1$ and $x_2$, and synthetic observation data was simulated using \eqref{eq:wobblyplate}, with $(x_1, x_2)$ drawn from $(\mathcal{U}(0.85, 1.6), \mathcal{U}(1.45, 1.85))$, creating the target density shown in Figures \ref{fig:wobblyplate}A, B. Figures \ref{fig:wobblyplate}A, B show the distribution of $\bm{y}$ obtained by push-forward with \eqref{eq:wobblyplate} of $x_1$ and $x_2$ sampled using Rejection, MCMC, c-GAN and r-GAN. Histograms of parameter samples for each method (shown in Figures \ref{fig:wobblyplate}D-G) demonstrate that all inference methods sample precisely from the `true' parameter distribution that was used to create the target data.

Adding stochasticity to the model, we created the synthetic target $\mathcal{Q}_Y$, shown in Figures \ref{fig:wobblyplate}I-J, by calculating \eqref{eq:wobblyplate} with $(x_1, x_2)$ drawn from $(\mathcal{U}(0.85, 1.6), \mathcal{U}(1.45, 1.85))$ and $(\epsilon_1, \epsilon_2)$ drawn from $(\mathcal{N}(0, 0.075^2), \mathcal{N}(0, 0.075^2))$, embedding noise in the locations at which the measurements were taken, as in \cite{butler2020data-consistent}. When sampling from $\mathcal{Q}_{X_g}$, the simulations were run with priors $(\mathcal{N}(0, 0.0825^2), \mathcal{N}(0, 0.0825^2))$ for $(\epsilon_1, \epsilon_2)$, reflecting a scenario where the true noise in the real system is not fully known. Figures \ref{fig:wobblyplate}I-J shows the distribution of $\bm{y}$ obtained by push-forward with \eqref{eq:wobblyplate} of $x_1$ and $x_2$ sampled using Rejection, MCMC, c-GAN and r-GAN. In this example, $\mathcal{Q}_{Y_g}$ for the rejection, MCMC-boosted rejection, and c-GAN methods are biased away from the target $\mathcal{Q}_Y$ (Figures \ref{fig:wobblyplate}I-J). This is because $(\epsilon_1, \epsilon_2)$ are only sampled a single time for each $(x_1, x_2)$, which induces a correlation between $\bm{x}$ and $\bm{\epsilon}$ in $\mathcal{Q}_X$, and the samples of $\bm{\epsilon}$ diverge from the true noise distribution. When samples from $\mathcal{Q}_X$ are then pushed forward through the MM, $(\epsilon_1, \epsilon_2)$ are resampled and $\mathcal{Q}_{Y_g}$ diverges from $\mathcal{Q}_Y$. Thus, these methods do not provide a solution that accurately estimates parameter density.

In the r-GAN framework, we were able to implement a more flexible network configuration that samples a parameterized model of noise during training. We incorporated a trainable parameter $\sigma$ (shown in Figure \ref{fig:wobblyplate}H) for the variance of a noise distribution $(\mathcal{N}(0, \sigma^2)$, constrained by an additional discriminator $D_{N}$ (Figure \ref{fig:wobblyplate}H) to produce samples close to the prior assumptions about the noise distribution. The noise parameter was able to gradually shift from the prior value of $0.0825^2$ towards the ground truth value of $0.075^2$ during training. This enabled model parameter samples from r-GAN to be highly concentrated in the ground truth region, indicating that the network was able to find a result more consistent with the observed data. A major benefit of framing SIP as a constrained optimization is the ability to incorporate additional constraints into the problem formulation, which is made straightforward by our use of a neural network graph as in r-GAN (Figure \ref{fig:rgan}).

\subsection{SIP for super-resolution imaging}

Our next experiment aimed to demonstrate SIP in a high-dimensional setting. We chose a super-resolution imaging scenario with the MNIST dataset for intuitive visualization of the results. The model $y = M(x)$ is average pooling over an input window of size 7x7 with stride 1 and without padding, which operates on a high resolution (HR) image as input $\bm{x}$ and outputs a blurred, low resolution (LR) image $\bm{y}$. Our dataset of images representing the prior distribution $\mathcal{P}_X$ was the MNIST dataset, and we passed MNIST images labelled with digit ``5'' through the model $\bm{y}=M(\bm{x})$ to obtain the target dataset $\mathcal{Q}_{Y}$.

\begin{figure}[h]
  \centering
  \includegraphics[width=1.0\linewidth]{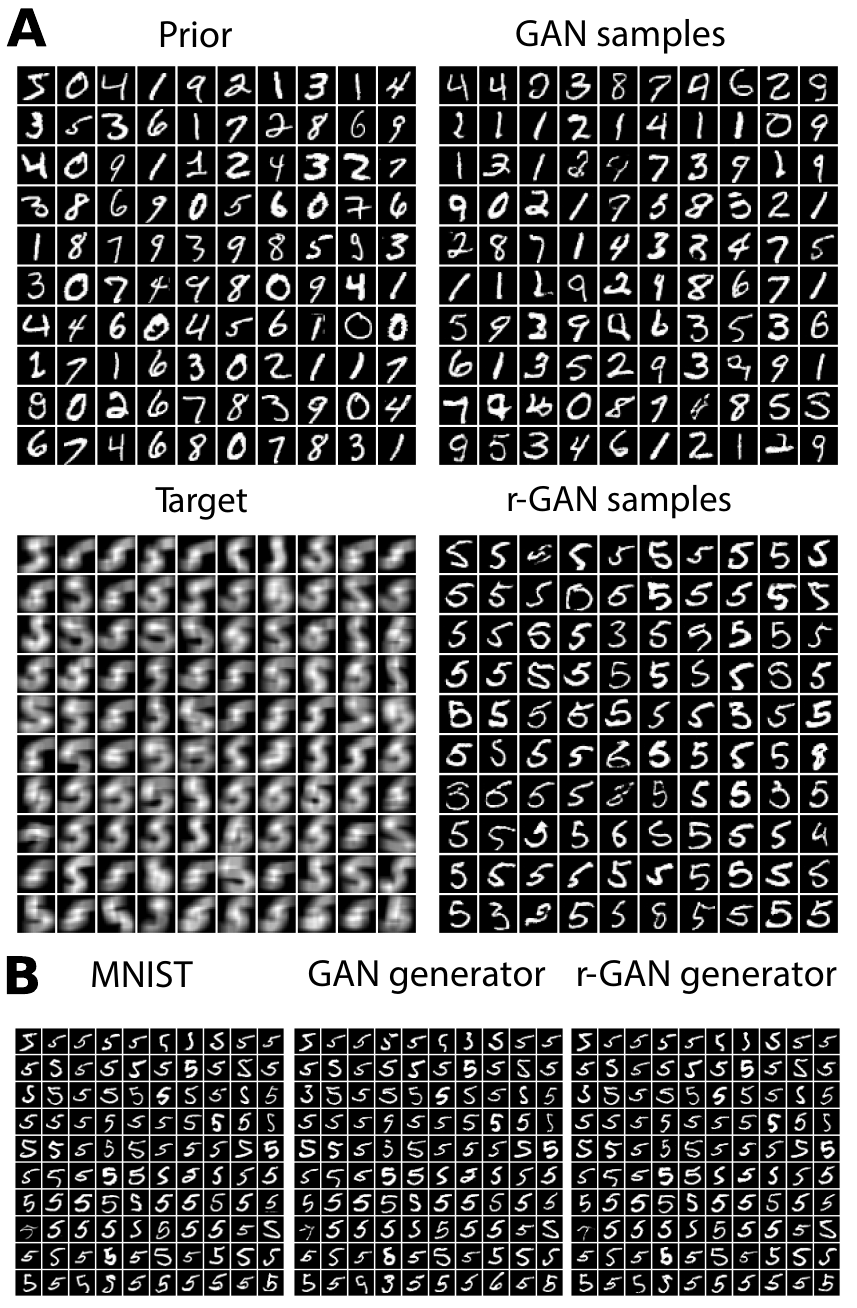}
  \caption{r-GAN for super-resolution imaging. \textbf{A.} MNIST dataset is used as a prior ($\mathcal{P}_X$) (top left). Target distribution $\mathcal{Q}_{Y}$ is obtained by average pooling of each image labeled ``5'' (bottom left). Initially, we trained a standard GAN and used weights for initialization of the r-GAN generator (samples in top right). After training, r-GAN generates samples from $\mathcal{Q}_{Xg}$, mostly images of ``5''s and some images that are close to ``5'' in the average pooling domain (bottom right). \textbf{B.} Finding high-resolution images with the PULSE model. Optimal images on the (d-1)-sphere of the latent space are found for blurred ``5'' digits from the MNIST dataset (left) using generators from GAN trained on the whole of MNIST (middle), and r-GAN trained on $\mathcal{Q}_{Y}$ with blurred ``5''s (right). Note that samples are shown in random order, and not aligned across panels.}
  \label{fig:MNIST}
\end{figure}

As an initial training stage in all r-GAN experiments, we trained the generator to reproduce the prior distribution by setting $w_{Y} = 0$ in \eqref{eq:LG}. For the super-resolution scenario, this amounts to training a standard GAN to produce MNIST images. Samples from $\mathcal{Q}_{X_g}$ after training are shown in Figure \ref{fig:MNIST}A, top right, and span the distribution of images from the full MNIST prior $\mathcal{P}_X$ (Figure \ref{fig:MNIST}A, top left). We added the loss of the second discriminator $D_{Y}$ by setting $w_{Y} = 1$ to train the complete r-GAN. Images generated by $G$ were passed through $\bm{y}=M(\bm{x})$ and compared with target samples, resulting in $\mathcal{Q}_{X_g}$ (Figure \ref{fig:MNIST}A, bottom right), which when passed through $M$, approximates $\mathcal{Q}_Y$ (Figure \ref{fig:MNIST}, bottom left). The samples consist of a variety of images that resemble those labelled ``5'', as well as occasional images that resemble those with different labels (most notably ``3''s, ``6''s and ``8''s), but which resemble LR ``5''s after applying $M$. Note that these images resembling alternatively labelled images are appropriate for the generator to sample, since, when blurred by $M$, they are indistinguishable from blurred ``5''s.

In other super-resolution imaging solutions, we observe that typically only a single image corresponding to a given LR image is sampled, as in the SRGAN \cite{ledig2017photo} and PULSE \cite{menon2020pulse} models. In PULSE, a GAN is first trained on the prior images. Then, the following objective is minimized: $min_{\bm{z} \in \mathcal{S}^{d-1}(\sqrt{d})} \| DS(G(\bm{z})) - I_{LR} \|$,
thus finding the latent vector $\bm{z}$ on the (d-1)-sphere with radius $\sqrt{d}$, where $d$ is the dimension of the latent space of the generator $G(\bm{z})$, $DS$ is the down-sampling function (our model $M$), and $I_{LR}$ is a LR image. We also used the PULSE model to find specific generated samples that closely match each image in $\mathcal{Q}_Y$. Using either the GAN generator (trained during the initial stage: $w_{Y} = 0$) or the r-GAN generator, PULSE found optimal latent space values that generated HR samples, which, when blurred, matched single LR samples from $\mathcal{Q}_Y$ (Figure \ref{fig:MNIST}B).

Evaluating generated samples in a high-dimensional space is challenging \citep{BORJI201941}. Here, we measured JS-divergence between both $\mathcal{P}_{X}$ and $\mathcal{Q}_{X_g}$, and between $\mathcal{P}_{Y}$ and $\mathcal{Q}_{Y_g}$, using a classifier trained to distinguish generated and real samples, shown in Table I. However, regularization of the classifier affects the calculated divergence metric, as the implicit density model defined by the classifier depends on its chosen regularization and metaparameters. Generated datasets (labelled in the left column of Table I) are compared against subsets of MNIST for different digits (0, 1, ..., 9) or the whole dataset (All), using either low (LR) or high (HR) resolutions. GAN and r-GAN labels indicate the generator used for samples. PULSE indicates sampling performed with PULSE model and a generator. We discuss the evaluations further, and provide details of the classifier, in Supplementary Information.

The low JS-divergence of 0.22 compared to other digits between the MNIST images labelled ``5'' and the samples from the r-GAN after applying $M$ to both (r-GAN (LR)) indicates that $Q_{Y_g}$ samples are close to the target $Q_{Y}$. The JS-divergence of 0.61 between the MNIST images labeled ``5'' and the samples from the r-GAN before applying $M$ (r-GAN (HR)) (i.e. $M$'s parameter space) can be compared with the values when sampling MNIST images with PULSE using either the GAN or r-GAN generator (0.73 and 0.71, respectively). This indicates that sampling from the r-GAN without attempting to select specific samples for each output of $M$ in the target dataset (i.e., treating superresolution imaging as a SIP) produces a distribution of HR images that is closer to the true distribution of ``5''s in MNIST than when sampling specific generator outputs that are optimal for each image in the dataset.

\begin{table*}[t]
\centering
\caption{}
\begin{tabular}{@{}llllllllllll@{}}
\toprule
                 & ``0'' & ``1'' & ``2'' & ``3'' & ``4'' & ``5'' & ``6'' & ``7'' & ``8'' & ``9'' & All \\ \midrule
r-GAN (LR)        & 0.96 & 0.98 & 0.96 & \textbf{0.86} & 0.97 & \textbf{0.22} & 0.93 & 0.98 & 0.92 & 0.96 &     \\
r-GAN (HR)        & 0.99 & 0.99 & 0.99 & 0.96 & 0.99 & \textbf{0.61} & 0.97 & 1.00 & 0.96 & 0.99 & \textbf{0.84}  \\
PULSE (GAN, LR)  & 0.98 & 0.98 & 0.96 & 0.89 & 0.97 & \textbf{0.07} & 0.94 & 0.97 & 0.91 & 0.94 &  \\
PULSE (GAN, HR)  & 1.00 & 1.00 & 0.99 & 0.97 & 0.99 & \textbf{0.73} & 0.98 & 0.99 & 0.97 & 0.98 & \textbf{0.91}    \\
PULSE (r-GAN, LR) & 0.98 & 0.98 & 0.96 & 0.89 & 0.98 & \textbf{0.07} & 0.95 & 0.98 & 0.94 & 0.96 &     \\
PULSE (r-GAN, HR) & 1.00 & 1.00 & 1.00 & 0.98 & 1.00 & \textbf{0.71} & 0.99 & 1.00 & 0.99 & 0.99 &     \\ \bottomrule
\end{tabular}
\end{table*}

\subsection{Additional SIP examples in Supplementary Information}

A fundamental benefit of the r-GAN formulation of SIP, as demonstrated in sections 3B and 3C, above, is the flexibility to reconfigure the framework to tailor it to entirely new types of SIP. Finally, we demonstrate a solution to a complex parameter inference problem that appears routinely in scientific disciplines using an intervention scenario, reflecting situations where recordings are made from a biological ensemble before and after a perturbation, such as administration of a drug to a tissue. In previous work, we applied such an intervention r-GAN to evaluation of the mechanism of action of a cardiac inotrope drug using a cardiac mechanics cellular model \cite{parikh2022OM}. Here, we include a test demonstration of this approach using the Rosenbrock function as the MM in the Supplementary Information.

To further validate the reliability of the r-GAN approach, we tested SIP using GANs for several other test functions, described in detail in the Supplementary Information. For example, we used the Rosenbrock function with a multi-modal target output distribution dependent on disjoint parameter regions. All methods (Rejection, MCMC and r-GAN) were able to accurately infer parameter samples coherent with the target distribution for a low-dimensional Rosenbrock function. In a high-dimensional Rosenbrock function example, the c-GAN struggled to constrain parameter samples to precisely match $\mathcal{Q}_Y$ and $\mathcal{Q}_{Y_g}$, due to the requirement that c-GAN learn the full conditional distribution. The MCMC and r-GAN methods performed well for this challenging test case, however.

For certain mechanistic models, a surrogate regression model approximating $\bm{y} = M(\bm{x})$ and trained on a dataset calculated from the original model is required to make an otherwise non-differentiable or costly simulation accessible to the r-GAN framework. Two additional SIPs, a piecewise smooth function (to test parameter inference in a discontinuous space) and an ODE model (a typical format for a mechanistic simulator) \citep{butler2018combining}, were solved using surrogate models in place of a direct solution to the model equations in the r-GAN. Parameters coherent with target outputs were sampled accurately for both the piecewise and ODE test functions by Rejection, MCMC, and r-GAN.

\section{Discussion}
\label{discussion}

Parameter estimation for MMs to fit ensembles of experimental observations in the domain of model output is a common task in biology. A typical problem is fitting MMs to biomarkers obtained from a set of cells with random variation of cell characteristics within the set. In practice, parameterizing MMs is a critical task when attempting to combine results of multiple experimental protocols in inference problems, and also for putting additional constraints into machine learning models in the form of an inductive bias provided by MMs. 

Probably the most difficult step in MM based analysis is parameter estimation. In the literature, investigators usually apply Bayesian methods to fit MMs to individual entities one-by-one, and then aggregate the found model parameters to build a density model in MM parameter space \citep{Harrod2021predictivemodeling}. The major drawback of this approach is that the prior on model parameters is rarely informative, and a uniform distribution within somewhat arbitrary ranges is often used. Thus, Bayesian analysis introduces unnecessary bias when there is little or no information available about MM parameters. Similar bias is also introduced in an approach termed `population of models' \citep{drovandi2016sampling}.

Data-consistent inversion in stochastic inverse problems provides a viable alternative to Bayesian parameter inference. In this approach, a density model of experimental observations is constructed, and the goal is to find a distribution of MM parameters that produces the target density. Although the prior also introduces a bias in the solution, the prior does not affect the distribution of model outputs. In this paper, we have presented new algorithms to solve stochastic inverse problems addressing limitations of the methods found in previous research. These new methods can be divided into two categories. The Rejection and r-GAN methods fall into the first category, which, similar to variational inference, optimizes distributions of model parameters by minimizing divergence between MM output distributions and target distributions. The distributions of MM parameters are represented either by sets of samples, as in the rejection methods, or by parametric density estimators, e.g., generator networks, as in the r-GAN. In both cases, samples are scored using discriminators in optimization algorithms. In the second category, instead of optimization, distributions of MM parameters for individual target samples are memorized by density estimators. We presented c-GAN as an example of this class of solution.

In the Rejection algorithm, we used Gaussian mixture models to estimate the densities of the sets of samples to compute a density ratio for the rejection step. As an alternative, we could employ the density ratio trick with a classifier, as in GANs, to minimize f-divergence between the target and MM output distributions. The MCMC-boosted Rejection algorithm aims to initialize the proposal distribution with a distribution closer to the solution, similar to the algorithm from \citep{lawson2018unlocking}. The main limitation of the Rejection algorithm is the prohibitively large number of initial samples from the prior that are required as dimensionality or problem complexity increases. Therefore, we looked for other algorithms in which particle sets are evolved within perturbation and resampling steps, similar to how sequential Monte Carlo methods are applied in Approximate Bayesian Computation. To construct such an algorithm, we reformulated the stochastic inverse problem into a constrained optimization problem and proposed to solve it in a series of unconstrained optimization steps. These steps could be implemented as a combination of perturbations according to a kernel, followed by application of the rejection algorithm with particle scores calculated from the sum of weighted discriminators \eqref{eq:coGAN}. As one potential method to solve SIP using stochastic gradient descent, we implemented r-GAN, a generative adversarial network with two discriminators.

A widely recognized limitation of GANs is that they are difficult to train, primarily caused by issues relating to mode collapse, thus limiting their adoption in general scientific applications. The stabilization method used here (see Supplementary Information for details) enabled stable training across all examples without extensive hyperparameter search. Mode collapse may still occur if a solution to SIP is in a low density region of the prior distribution, but one potential solution to this is to combine an explicit density estimator (e.g., normalizing flow networks) with a classifier to apply boosting for training the prior discriminator \citep{cranko2019boosted}. We anticipate such limitations of GANs can in the future be mitigated by more advanced training techniques and GAN configurations. 

One requirement of r-GAN is that the mechanistic model must be differentiable to allow backpropagation of $L_{D_Y}$ to $G$. However, mechanistic simulators are often numerical solutions of differential equations, and cannot be directly incorporated into a deep network. In two of our test examples, a differentiable surrogate model was  trained on samples from $\mathcal{P}_X$ paired with their outputs of $M$ and used in place of the mechanistic model (see Supplementary Information). This approach can introduce error if the target region of interest is undersampled during surrogate model training. Alternatively, instead of using a neural network for the GAN generator, distributions of model parameters can be represented by a set of particles, and the objective of \eqref{eq:coGAN} can be optimized by applying perturbations and resampling to the particle set. Parallels can be made to sequential Monte Carlo sampling methods \citep{doucet2001sequential, DelMoral2021smc} used in ABC \citep{Sisson2018}, a direction we are currently actively exploring. Active learning approaches, such as those used for sequential refinement of conditional density model training data from SBI studies \citep{lueckmann2019likelihood, goncalves2020training} can also be applied to iteratively refine a surrogate model. 

Fast and accurate solutions to SIPs are critical in any physical science domain where simulations are routinely used to model populations of observations. In the domain of computational biology, specifically the field of cardiac mechanics, we have used the intervention methods presented here to perform an evaluation of the mechanism of action of various cardiac inotropes, applying our methods to unloaded contraction cell data and a cardiac mechanics cellular model \cite{parikh2022OM}. The methods are also being actively explored for analysis of cardiac and neuronal electrophysiology data as in \cite{smirnov2020genetic} and \cite{rumbell2019dimensions}.

The r-GAN extends GANs' applications beyond generative modeling and variational inference \citep{huszar2017variational}. Our first experiment showed that this method solves SIP for systems of equations with accuracy equaling rejection methods. We also formulated three SIP scenarios that cannot currently be solved using existing methods. First, we demonstrated an improvement in predictions for SIP with stochastic models by incorporating a network for parameterizing noise into the optimization. We showed how alternative methods fail at this task. Second, we formulated a super-resolution imaging problem as a SIP, demonstrating the use of r-GAN with convolutional networks. The rejection sampling algorithm would be prohibitively slow to converge when tackling high-dimensional problems of this type. Instead, using r-GAN we were able to initialize the generator to the prior by means of conventional GAN training, and then perform gradient descent during optimization towards the SIP solution. Our method is therefore the only method for solving problems of this type. Third, in the Supplementary Information, we introduce one of many possible novel configurations of r-GAN that solves SIP within an intervention experiment, as demonstrated previously for a cardiac inotrope \citep{parikh2022OM}. The success of these test experiments leads us to propose that approaches based on our constrained optimization formulation are currently the only path to solving a broad family of related SIPs with equally broad applications. 

\bibliographystyle{IEEEtran}
\bibliography{main}

\end{document}


\title{Novel and flexible parameter estimation methods for data-consistent inversion in mechanistic modeling}

\author{
    Timothy Rumbell\textsuperscript{\rm 1},
    Jaimit Parikh\textsuperscript{\rm 1}, 
    James Kozloski\textsuperscript{\rm 1},
    Viatcheslav Gurev\textsuperscript{\rm 1*}
\thanks{
        \textsuperscript{\rm 1} IBM Research, Hybrid Biological-AI Modeling
        
        \textsuperscript{\rm *} Corresponding author: vgurev@us.ibm.com}
}

\maketitle

\section*{Supplementary Information}

%

\section{Stochastic Inverse Problems (SIP) in an intervention scenario using r-GAN extensions}
\label{sec:interventionSIP}

A common research scenario for mechanistic modeling, especially in biology, involves recordings of ensembles under different conditions. For example, to find the effect of a candidate drug compound, characteristics of two sets of isolated cells may be recorded, one under control conditions and the other under the effects of the drug. The problem of finding input parameters of a model for multiple conditions distinguished by some factor (e.g., drug action, age, disease state, etc.) can be framed as an intervention problem, in which a subset of model parameters (potentially known to be related to the mechanism of action of the intervention) are allowed to vary across conditions, while other model parameters remain unchanged.

For an example of parameter inference in an intervention scenario, we set the goal of simultaneously inferring MM parameters for two sets of observations: the first under a `control' condition, and the second under a `drug' condition. We adapted the r-GAN architecture to solve SIP in this scenario. Here we denote by $\bm{x}_c\sim \mathcal{Q}_{X_c}$, $\bm{x}_d\sim \mathcal{Q}_{X_d}$ samples of model input parameters for the control population and treatment populations under the drug, respectively. Our goal was to evaluate distributions $\mathcal{Q}_{X_c}$ and $\mathcal{Q}_{X_d}$ given distributions of observations $\mathcal{Q}_{Y_c}$ and $\mathcal{Q}_{Y_d}$ for the control and treatment populations. Note that we consider the situation where we do not have pairwise observations for each object under both control and drug conditions. We address this type of scenario because it is prevalent in healthcare and life sciences domains, such as randomized clinical trials. Inferring model input parameters in the less common situation where pairwise observations are available can be solved more simply. 

To proceed, we define a joint probability distribution between $X_c$ and $X_d$ with marginals $\mathcal{Q}_{X_c}$ and $\mathcal{Q}_{X_d}$. Interventions rarely affect the whole set of model input  parameters. Often, input parameter vectors can be split into the components $\bm{x}_s$ that are not affected by the drug (shared parameters) and components $\bar{\bm{x}}_c$, $\bar{\bm{x}}_d$ forming two vectors of input parameters $\bm{x}_c = [\bm{x}_s, \bar{\bm{x}}_c]$, ${\bm{x}}_d = [\bm{x}_s, \bar{\bm{x}}_d]$ for the control and treatment groups, respectively. The split results in the factorization $q_{\bar{X_c},\bar{X_d}|X_s}(\bar{\bm{x}}_c, \bar{\bm{x}}_d | \bm{x}_s) = q_{\bar{X_c}|X_s}(\bar{\bm{x}}_c| \bm{x}_s)q_{\bar{X_d}|X_s}(\bar{\bm{x}_d}| \bm{x}_s)$. This problem cannot be solved independently for the two populations. However, the extension of r-GAN to accommodate this problem is straightforward:

\begin{equation} \label{eq:tgan}
\begin{aligned}
& \text{given}\; & & \mathcal{P}_{X_c},\;  \mathcal{P}_{X_d},\; \mathcal{Q}_{Y_{c}},\; \mathcal{Q}_{Y_{d}},\; M \\
& \underset{\theta_1,\theta_2,\theta_3}{\text{minimize}}
& & D_f(\mathcal{P}_{X_c} || \mathcal{Q}_{X_{g,c}}) + D_f(\mathcal{P}_{X_d} || \mathcal{Q}_{X_{g,d}})  \\
& \text{subject to}
& & supp(X_{g,c})\subseteq supp(X_c),\;
\\ & & &supp(X_{g,d})\subseteq supp(X_d),\;
\\& & &D_f(\mathcal{Q}_{Y_c} || \mathcal{Q}_{Y_{g,c}}) =0,\; D_f(\mathcal{Q}_{Y_d} ||  \mathcal{Q}_{Y_{g,d}}) = 0 \\
& \text{where}\; & & [\bm{z}_s, \bm{z}_c, \bm{z}_d] \sim \mathcal{P}_Z,\;
\\& & &\bm{x}_s = G_{\theta_1}(\bm{z}_s),\;\bar{\bm{x}}_c =  G_{\theta_2}(\bm{z}_c, \bm{z}_s),\;
\\& & &\bar{\bm{x}}_d =  G_{\theta_3}(\bm{z}_d, \bm{z}_s),
\\& & &\bm{x}_c = [\bm{x}_s, \bar{\bm{x}}_c],\; \bm{x}_d = [\bm{x}_s, \bar{\bm{x}}_d],
\\& & &\bm{x}_c\sim \mathcal{Q}_{X_{g,c}},\; \bm{x}_d\sim \mathcal{Q}_{X_{g,d}},\\& & & M(\bm{x}_c)\sim \mathcal{Q}_{Y_{g,c}},\; M(\bm{x}_d)\sim \mathcal{Q}_{Y_{g,d}}. \\
\end{aligned}
\end{equation}

\begin{figure*}[h!]
  \centering
  \includegraphics[width=0.8\textwidth]{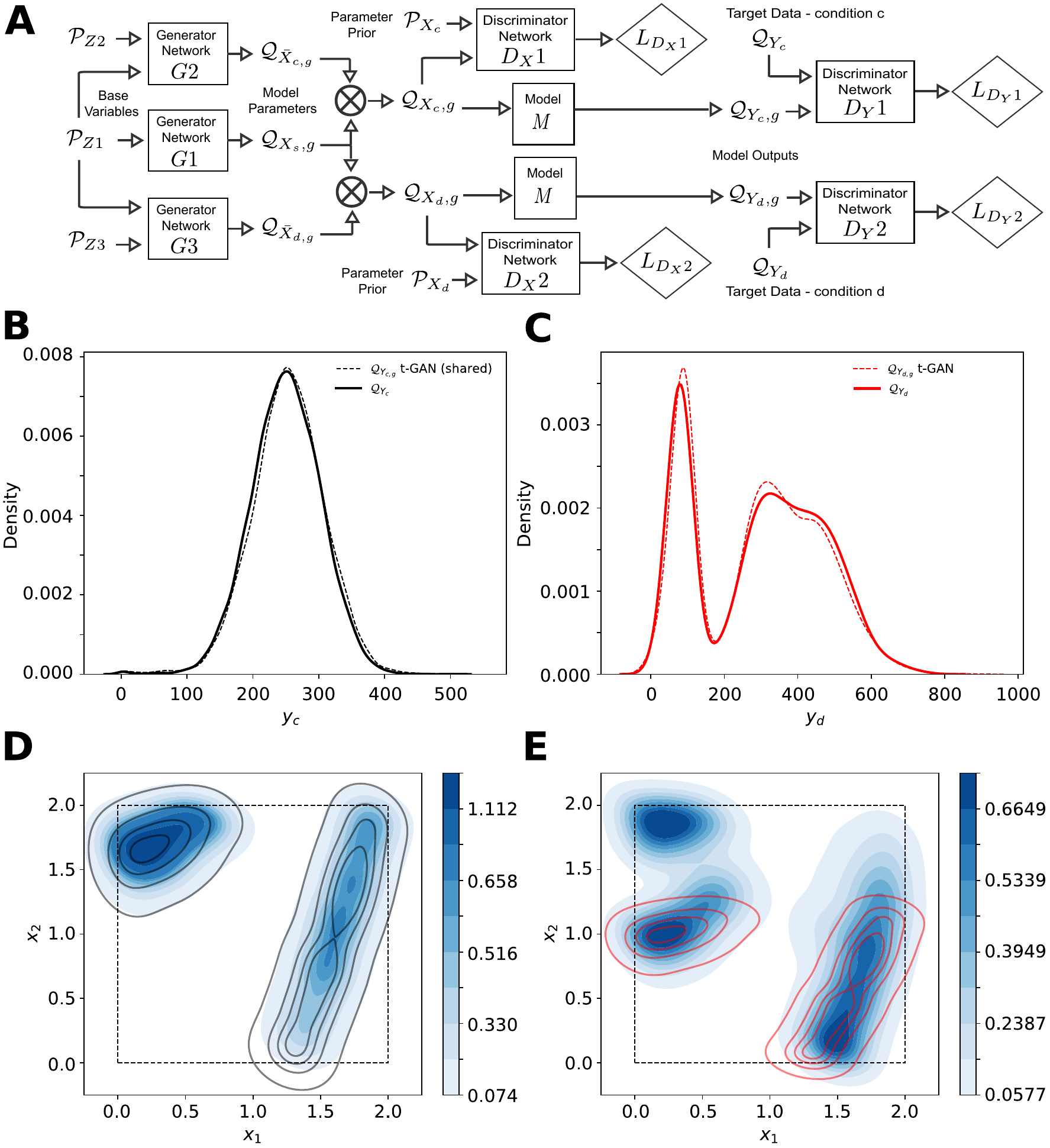}
 \caption{\textbf{A} r-GAN for parameter inference in an intervention scenario. Here the intervention is applied to a subset of parameters $\bar{X}$, while a generator $G1$ samples unaffected, shared parameters $X_{s}$. Two additional generators, $G1$ and $G2$, respectively sample $\bar{X}$ under the intervention and non-intervention conditions $d$ and $c$. Two simulations produce model outputs from parameters generated for each condition. This r-GAN has four discriminators contributing to the generators' losses in weighted sums. \textbf{B}. KDEs of target distribution under control conditions $\mathcal{Q}_{Y_{c}}$ (black solid) line and generated (inferred) output distribution $\mathcal{Q}_{Y_{c,g}}$ (dashed line). \textbf{C}. KDEs of target distribution after intervention, $\mathcal{Q}_{Y_{d}}$ (red solid line) and the inferred output distribution $\mathcal{Q}_{Y_{d,g}}$ (dashed line). \textbf{D}. Joint distribution of model input parameters for control observations for ground truth parameters $\mathcal{G}_{X_{c}}$ used to generate the observations (black contour lines), and inferred parameters via r-GAN (contour map in blue). \textbf{E}. As in \textbf{D}, but after intervention for ground truth parameters $\mathcal{G}_{X_{d}}$ (red contour lines), and inferred parameters via r-GAN (contour map in blue).}
  \label{fig:tgan}
\end{figure*}

The graph of the intervention r-GAN with shared parameters is shown in Figure \ref{fig:tgan}A. Through a combination of three generators with both shared ($Z1$) and unshared ($Z2$, $Z3$) base variables, experimental information is incorporated into the structure of the GAN itself. Four discriminators are used to provide the generators' losses in a weighted sum as in \eqref{eq:LG} below: two ensure that the mechanistic model outputs match both control and drug observations, and two maximise the overlap between sampled parameter sets and the parameter priors. 

To demonstrate the intervention r-GAN configuration, we used a synthetic dataset following the intervention scenario. A 2-dimensional Rosenbrock function,
\begin{equation} \label{eq:ros_2d}
    M(\bm{x}) = (a-x_1)^2 + b(x_2-x_1^2)^2,
\end{equation}
 was used as the mechanistic model. We generated samples of observations with distribution $\mathcal{Q}_{Y_c}$, corresponding to the control condition, from $\mathcal{N}(250,50^2)$, shown in Figure \ref{fig:tgan}B, solid black line. The ground-truth distribution of input parameters $\mathcal{G}_{X_{c}}$ coherent to $\mathcal{Q}_{Y_c}$ is shown in Figure \ref{fig:tgan}D as black contour lines. Next, we sampled from the distribution of ground truth input parameters $\mathcal{G}_{X_{c}}$ and applied linear scaling to the $x_2$ parameter according to $x_{2,d} = 0.6x_{2,c}$, to generate ground truth input parameters for observations under the intervention (drug) condition. Note that the input parameter $x_1$ is considered to be the shared input parameter, $x_s$, which is known to be unaffected by the intervention. The ground-truth distribution of input parameters after intervention $\mathcal{G}_{X_d}$ is shown in Figure \ref{fig:tgan}E as red contour lines. Equation \ref{eq:ros_2d} was run with $\mathcal{G}_{X_d}$ as inputs to obtain the intervention target output distribution $\mathcal{Q}_{Y_d}$, shown in Figure \ref{fig:tgan}C, solid red line. We reemphasize that we do not have pairwise data for each object under control and after intervention conditions, i.e., we do not have information on the joint distribution of observations across the two sets, but only the marginal distributions $Q_{Y_c}$ and $Q_{Y_d}$ of the separate observations.

We used the r-GAN with shared variables \eqref{eq:tgan} (Figure \ref{fig:tgan}A) to infer model input parameters coherent with the observations $\mathcal{Q}_{Y_c}$ and $\mathcal{Q}_{Y_d}$. The distributions of inferred input parameters under control and intervention conditions are shown in Figures \ref{fig:tgan}D and \ref{fig:tgan}E by the blue contour maps. The generated distributions of input parameters resulted in the output observation distributions shown by the dotted density lines in Figures \ref{fig:tgan}B and \ref{fig:tgan}C, which closely match both target distributions. In this scenario, there were not sufficient prior constraints for the inferred parameter distribution after intervention (Figure \ref{fig:tgan}E) to precisely match the ground truth distribution (i.e. the true effect of the intervention on model parameters) as demonstrated by the blue shaded regions outside of the red contour lines. Note, however, that by targeting a uniform distribution in parameter space, and while being constrained to match the observations in output space, the r-GAN inferred a wider range of possible effects of the intervention than were present in the ground truth effects, i.e., the intervention r-GAN result contains the actual intervention effect as one of multiple possible coherent solutions.

\begin{figure*}[h!]
  \centering
  \includegraphics[width=0.8\textwidth]{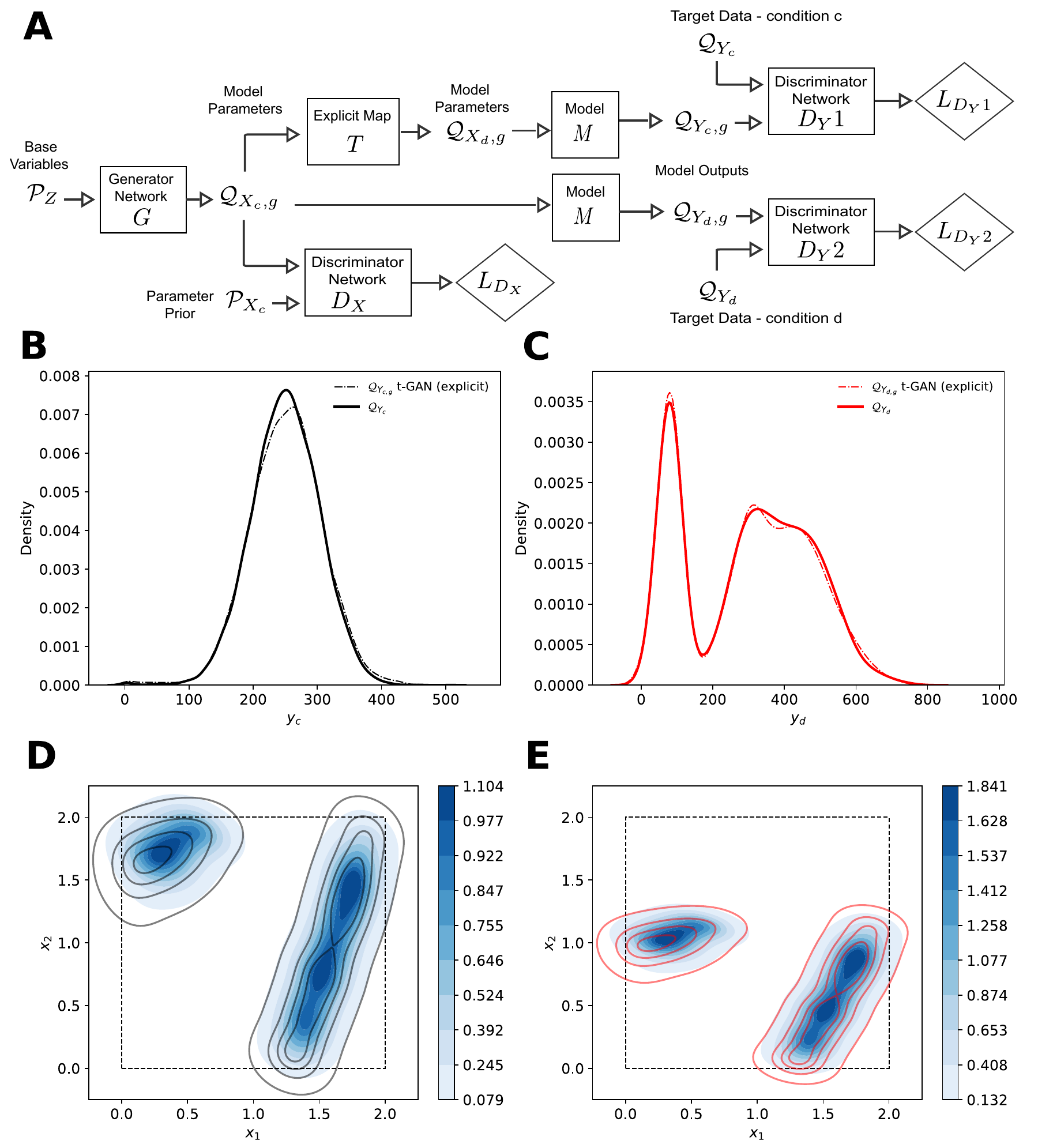}
 \caption{r-GAN modified for an explicitly known intervention scenario. \textbf{A.} An explicit map $T$ can be applied to generated parameters under the control condition $\mathcal{Q}_{X_c,g}$ to produce parameters under the perturbed condition $\mathcal{Q}_{X_d,g}$. Two mechanistic simulations produce model outputs from parameters generated for each condition. This r-GAN has three discriminators contributing to the generators' losses in weighted sums. The r-GAN enforces the equality of both $\mathcal{Q}_{Y_c}$ with $\mathcal{Q}_{Y_c,g}$ and $\mathcal{Q}_{Y_d}$ with $\mathcal{Q}_{Y_d,g}$, while maximizing the overlap between $\mathcal{P}_{X_c}$ and $\mathcal{Q}_{X_c,g}$. \textbf{B}. KDEs of the target distribution under control conditions $\mathcal{Q}_{Y_{c}}$ (black solid) line and the generated (inferred) output distribution $\mathcal{Q}_{Y_{c,g}}$ via r-GAN with explicit mapping (dashed dotted line). \textbf{C}. KDEs of the target distribution after intervention, $\mathcal{Q}_{Y_{d}}$ (red solid line) and the inferred output distribution $\mathcal{Q}_{Y_{d,g}}$ via the r-GAN with explicit mapping (dashed dotted line). \textbf{D}. Joint distribution of the model input parameters inferred via r-GAN with explicit mapping (contour map in blue) for the control observations with distribution $\mathcal{Q}_{Y_{c}}$ (black contour lines). \textbf{E}. As in \textbf{D}., but for joint distribution after intervention (contour map in blue) for $\mathcal{Q}_{Y_{d}}$ (red contour lines).}
  \label{fig:tgan_exp}
\end{figure*}

\textbf{Explicitly known deterministic map.} Next, we modifed r-GAN to demonstrate its flexibility to adapt to a second scenario, i.e., an intervention with a known effect. We simulated the intervention with an explicit deterministic map $\bm{x}_d = T(\bm{x}_c)$, and used this to infer the ensemble of MM parameters consistent with data observed before and after the intervention. This configuration would be useful for a scenario where the effect of the perturbation is fully understood. For example, a drug with known effect on a specific cellular membrane protein may be employed to test the response of a cell in an experiment. A suitable r-GAN to solve this intervention SIP (Figure \ref{fig:tgan_exp}A) is then:

\begin{equation} \label{eq:cGANex2}
\begin{aligned}
& \text{given}\; & & \mathcal{P}_{X_c},\; \mathcal{Q}_{Y_{c}},\; \mathcal{Q}_{Y_{d}},\; M \\
& \underset{\theta}{\text{minimize}}
& & D(\mathcal{P}_{X_c}||\mathcal{Q}_{X_{g,c}}) \\
& \text{subject to}
& & supp(X_{g,c})\subseteq supp(X_{c}),\;
\\& & & D(\mathcal{Q}_{Y_c}||\mathcal{Q}_{Y_{g,c}}) =0,\; D(\mathcal{Q}_{Y_d}|| \mathcal{Q}_{Y_{g,d}}) = 0 \\
& \text{where}\; & & \bm{z}\sim \mathcal{P}_Z,\; \bm{x}_c = G_\theta(\bm{z}), 
\\& & &\bm{x}_c\sim \mathcal{Q}_{X_{g,c}},\; \bm{x}_d = T(\bm{x}_c),\\& & & M(\bm{x}_c)\sim \mathcal{Q}_{Y_{g,c}},\; M(\bm{x}_d)\sim \mathcal{Q}_{Y_{g,d}}.
\end{aligned}
\end{equation}

We used the r-GAN with explicit map \eqref{eq:cGANex2} (Figure \ref{fig:tgan_exp}A) to infer model input parameters coherent with the observations $\mathcal{Q}_{Y_c}$ and $\mathcal{Q}_{Y_d}$ (Figures \ref{fig:tgan_exp}B and \ref{fig:tgan_exp}C), which were the same as in the intervention with shared variables. The r-GAN with explicit deterministic map produced distributions of input parameters shown in Figures \ref{fig:tgan_exp}D and \ref{fig:tgan_exp}E by the blue contour maps, which closely match the ground truth distribution of input parameters. The output distribution of the function, corresponding to the generated input parameters, is shown by the dashed-dotted density lines in Figures \ref{fig:tgan_exp}B and \ref{fig:tgan_exp}C.

These two cases represent two extremes of the knowledge that may be available about the intervention's effect. In the shared variables case, independent input parameters imposed the weakest possible constraint on the joint distribution. We emphasize that despite the factorization, the joint density derived from our novel methods does not necessarily assume that no correlation between $X_c$ and $X_d$ exists. It is possible to construct an infinite number of joint distributions for input parameters $x_c$ and $x_d$, each of which yields exactly same marginal distributions $\mathcal{Q}_{X_{c}}$, $\mathcal{Q}_{X_{d}}$ of the input parameters and generates the same distributions of output observations $\mathcal{Q}_{Y_{c}}$ and $\mathcal{Q}_{Y_{d}}$. The factorization of joint density only implies the degree of uncertainty about the joint distribution, with independent treatment of the two populations. The joint distribution of input parameters $\mathcal{Q}_{X_{c}}$, $\mathcal{Q}_{X_{d}}$ in the final solution can then be chosen based on some additional criteria, e.g., by solving the optimal transport problem \citep{peyre2019computational} on inferred marginals $\mathcal{Q}_{X_{c}}$, $\mathcal{Q}_{X_{d}}$. 

To construct the second extreme case, we assumed a strong constraint, i.e., we treated the relationship between input parameters in the two groups as a known deterministic map. In this case, the r-GAN correctly inferred the ground truth parameter distributions (Figures \ref{fig:tgan_exp}D and \ref{fig:tgan_exp}E). For simplicity of the presentation, we provided only these two extreme examples, leaving other configurations for future work. For example, it is also possible to construct joint distributions that lie between the two extremes using an r-GAN configuration that accounts for smooth responses to the intervention. Smooth response might be implemented using different configurations of generator networks with additional regularization, e.g, enforcing Lipschitz continuity in neural networks \citep{miyato2018spectral, gouk2018regularisation}.


\section{Test function examples}

We tested the Rejection algorithm using Gaussian Mixture Models (GMMs), with and without Markov chain Monte Carlo (MCMC) for sample initialization, along with c-GAN and r-GAN, on several example test functions, in addition to those shown in Figures 2 and 3 of the main manuscript. The examples comprised both two-dimensional and high-dimensional Rosenbrock functions, a two-dimensional piecewise discontinuous function \citep{butler2018combining}, and an ordinary differential equation model with two inputs \citep{breidt2011measure}.

\subsection{GMM and MCMC methods}

To compare our methods, we performed rejection sampling using algorithm 2 in the main manuscript. Samples for the rejection sampling were initialized using either samples from the prior distribution (notated as `Rejection' or `Rej') or using a MCMC method implemented using TensorFlow libraries (notated as `MCMC'). To calculate log-density under target distributions in all methods, we fit a Gaussian mixture model to the target samples from $\mathcal{P}_Y$, with the same samples used as the target in all GAN methods. For the MCMC initialization step, we initialized workers by sampling from $\mathcal{P}_{X}$, and then used the No U-Turn Sampler \citep{hoffman2014no} to generate initial proposals. 

\subsection{Two-dimensional Rosenbrock function}

\begin{figure*}[htbp]
  \centering
    \includegraphics[width=0.8\textwidth]{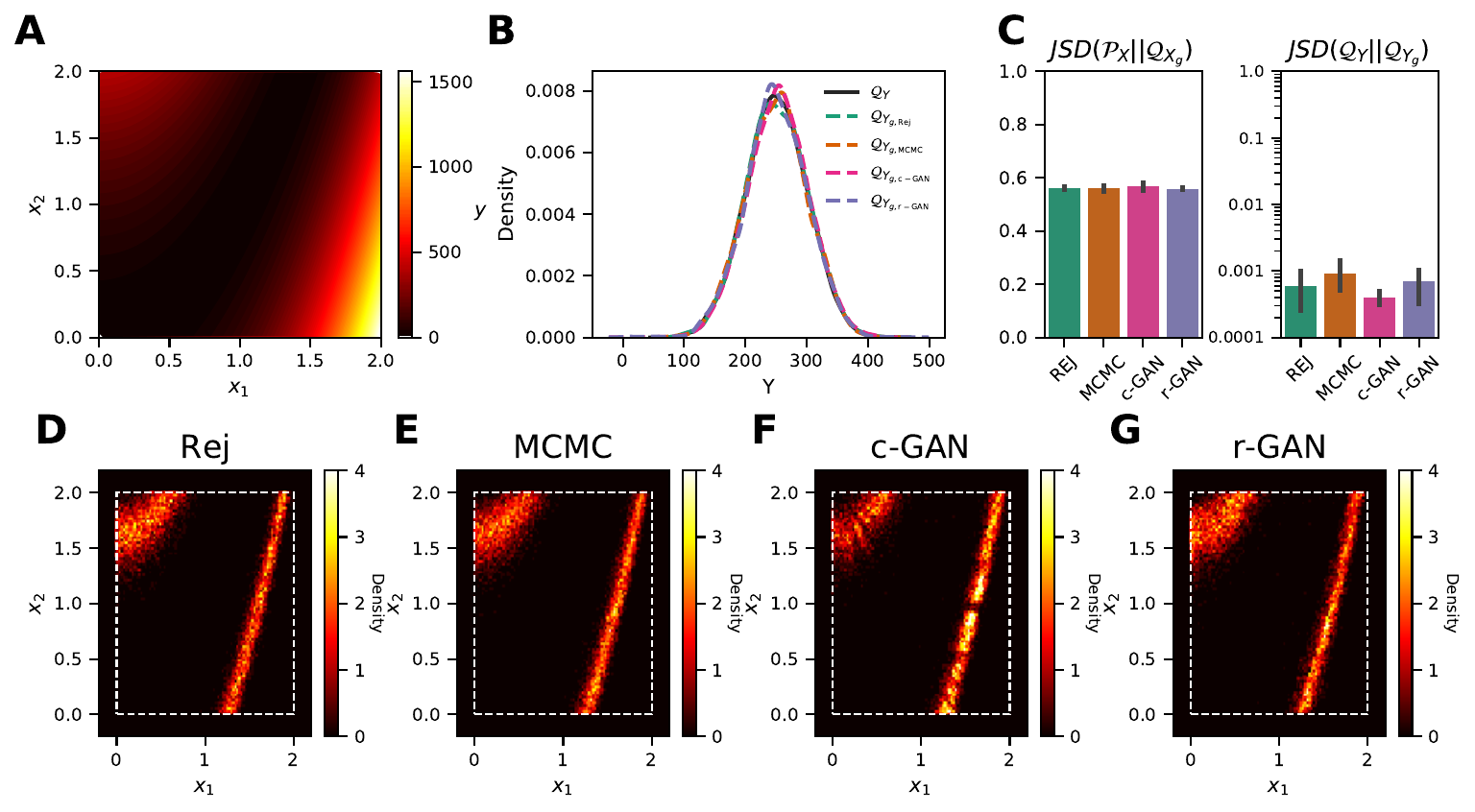}
  \caption{Two-dimensional Rosenbrock function test. \textbf{A}. Heat map of $y$ over the prior. \textbf{B}. Kernel density estimation (KDE) of the desired target output distribution $\mathcal{Q}_Y$ (black) and the generated (inferred) output distributions $\mathcal{Q}_{Y_g}$ using Rejection, MCMC, c-GAN and r-GAN (in green, orange, pink and purple, respectively). \textbf{C}. Left: estimated JS-divergence between samples from $\mathcal{P}_X$ and $\mathcal{Q}_{X}$ and Right: estimated JS-divergence between samples from $\mathcal{Q}_Y$ and $\mathcal{Q}_{Y_g}$ for all methods. \textbf{D-F}. 2D histograms of $\mathcal{Q}_X$ for Rejection (\textbf{D}), MCMC (\textbf{E}), c-GAN (\textbf{F}), and r-GAN (\textbf{G}). The dashed rectangle denotes the bounds set by the prior $\mathcal{P}_X$.}
  \label{fig:rosenbrock2d}
\end{figure*}

We tested the SIP methods in a scenario in which a non-linear model is clearly non-invertible, in that multiple disjoint modes in parameter space exist, each capable of producing the full distribution of target model outputs. Using the Rosenbrock function of two input parameters,

\begin{equation} \label{eq:ros}
    M(\bm{x}) = (a-x_1)^2 + b(x_2-x_1^2)^2,
\end{equation}
with $a = 1$, $b = 100$ (Figure \ref{fig:rosenbrock2d}A), over the prior range ($x_1, x_2 \sim \mathcal{U}(0, 2)$), the SIP methods were used to infer the joint distribution of input parameters coherent with target samples from a distribution $\mathcal{Q}_Y$ of $\mathcal{N}(250,50^2)$ (Figure \ref{fig:rosenbrock2d}B).

Calculating outputs according to \eqref{eq:ros} for the inferred input parameter samples $\mathcal{Q}_{X_g}$ generated by each method resulted in the model output distributions $\mathcal{Q}_{Y_g}$ shown in Figure \ref{fig:rosenbrock2d}B. The generated output distribution almost perfectly matches the desired target distribution. JS-divergence estimates (Figure \ref{fig:rosenbrock2d}C) show that all four methods perform similarly in sampling parameters coherent with the target samples. Figure \ref{fig:rosenbrock2d}D-G show histograms of $\mathcal{Q}_{X_g}$ for each method.

\subsection{High dimensional Rosenbrock function}
\label{sec:roshd}

To mimic the complexity of most biophysical models, we also considered a Rosenbrock function with multidimensional inputs,  

\begin{equation} \label{eq:ros_hd}
    f(\bm{x}) = \sum_{i=1}^{N-1}[b (x_{i+1}-x_i^2)^2 + (a-x_i)^2],
\end{equation}
with $a = 1$, $b = 100$, and the dimension $N$ set to 8. To generate a model $M$ with a vector of outputs $\bm{y}$ rather than a scalar, we performed 5 randomly chosen permutations of the coordinates $\{x_i\}$ in \eqref{eq:ros_hd}, yielding the 5 dimensional output vector (i.e., the dimensions of $X$ and $Y$ were 8 and 5, respectively):

\begin{equation} \label{eq:ros_hd_perm}
    M(\bm{x}) = [f(\bm{x}^1), f(\bm{x}^2), \dots, f(\bm{x}^5)],
\end{equation}
where $\bm{x}^i$ comprise permutations of the vector $\bm{x}$. Similar to the Rosenbrock function of two input parameters, we considered a uniformly distributed prior for the high dimensional model $x_i \sim \mathcal{U} (0, 2)$.

\begin{figure*}[htbp]
  \centering
    \includegraphics[width=0.8\textwidth]{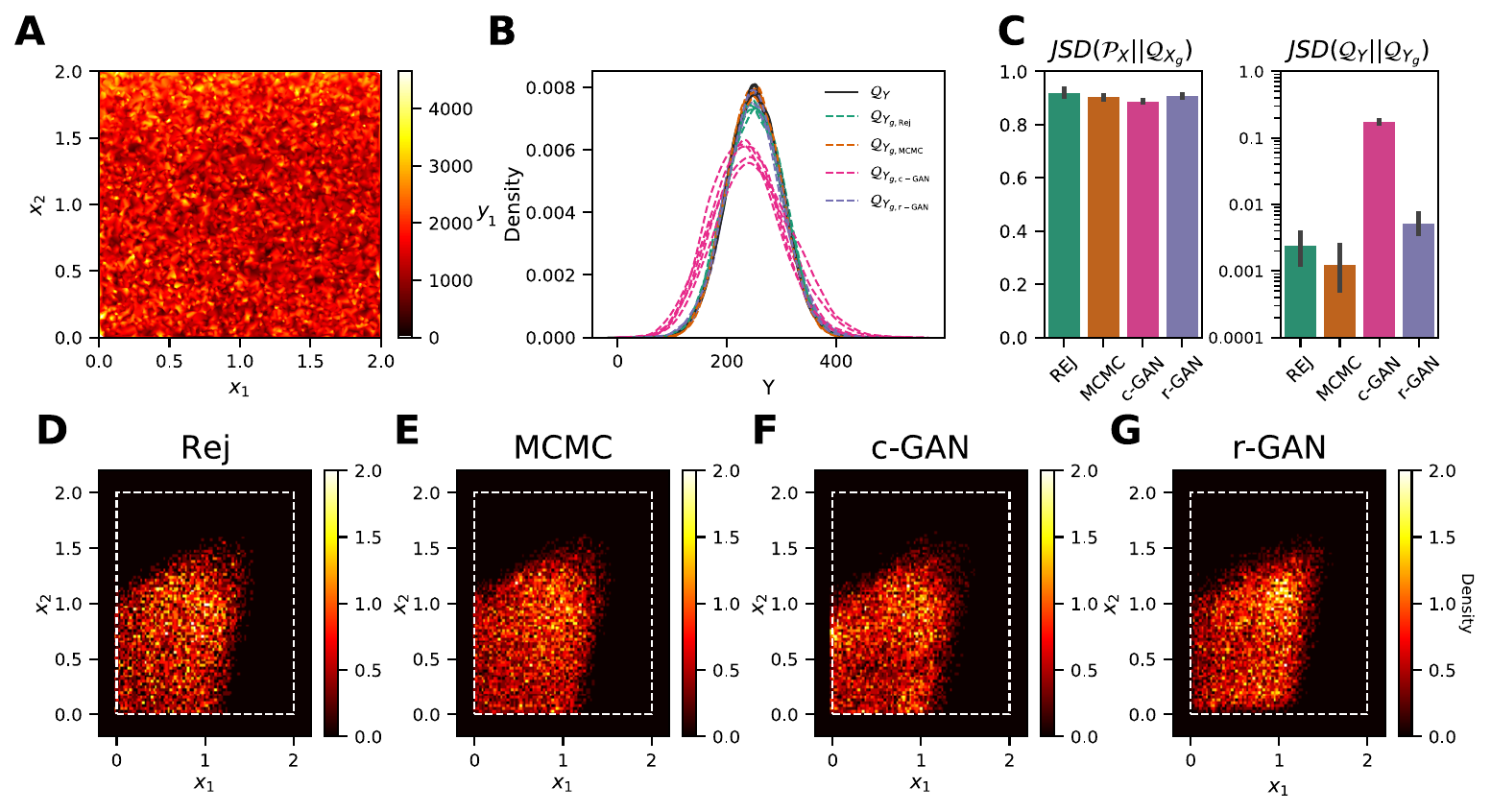}
  \caption{High-dimensional Rosenbrock function test. \textbf{A}. Heat map of $y_1$ over the marginal prior for $x_1$ and $x_2$. Note, only 2 dimensions of $x$ and 1 dimension of $y$ are displayed for visualization \textbf{B}. Marginal KDEs of the desired target output distribution $\mathcal{Q}_Y$ (black) and the generated (inferred) output distributions $\mathcal{Q}_{Y_g}$ using Rejection, MCMC, c-GAN and r-GAN (in green, orange, pink and purple, respectively). Multiple lines show marginals for all $5$ dimensions of $\bm{y}$ \textbf{C}. Left: estimated JS-divergence between samples from $\mathcal{P}_X$ and $\mathcal{Q}_{X}$ and Right: estimated JS-divergence between samples from $\mathcal{Q}_Y$ and $\mathcal{Q}_{Y_g}$ for all methods. \textbf{D-F}. 2D marginal histograms of $x_1$ and $x_2$ from $\mathcal{Q}_X$ for Rejection (\textbf{D}), MCMC (\textbf{E}), c-GAN (\textbf{F}), and r-GAN (\textbf{G}) for $x_1$ and $x_2$. The dashed rectangle denotes the bounds set by the prior $\mathcal{P}_X$. Note, only 2 dimensions of $x$ are displayed for visualization.}
  \label{fig:rosenbrockhd}
\end{figure*}

We applied the SIP methods to infer parameters for this 8-dimensional Rosenbrock function according to \eqref{eq:ros_hd}, with 5-dimensional output according to \eqref{eq:ros_hd_perm}. The target output distribution $\mathcal{Q}_Y$ was a multivariate normal distribution with means $\mu_i = 250$, $i = 1,2,\dots, 5$ and diagonal covariance matrix with standard deviation of each individual features $\sigma_{Y_{i}} = 50$, $i = 1,2,\dots, 5$. Figure \ref{fig:rosenbrockhd}C shows estimated JS-divergences between $\mathcal{P}_X$ and $\mathcal{Q}_X$ and between $\mathcal{Q}_Y$ and $\mathcal{Q}_{Y_g}$. Here, unlike the other methods, to capture the inverse surrogate of the model, the c-GAN needs to learn the multidimensional output function over the entire support of the prior, then perform amortized inference, resulting in comparatively poor performance of the c-GAN in this scenario. Rejection, MCMC and r-GAN had similar performance at sampling from the coherent distribution.

\subsection{A piecewise smooth function}

Another parameter inference test in \citep{butler2018combining} used a piecewise smooth function, demonstrating model parameter inference within a disconnected and compact region of a discontinuous function. The mechanistic model was represented by the equation:

\begin{equation} \label{eq:piecewise}
  \bm{y} = M(\bm{x}) =
    \begin{cases}
      q_1(\bm{x}) - 2 & 3x_1 + 2x_2 \geq 0 \text{ and } \\ 
                      & -x_1 + 0.3 x_2 < 0  \\
      2 q_2(\bm{x}) & x_1 + 2x_2 \geq 0 \text{ and } \\ 
                      & -x_1 + 0.3 x_2 \geq 0 \\
      2 q_1(\bm{x}) + 4 & (x_1 + 1)^2 + (x_2 + 1)^2 < 0.95^2 \\
      q_1(\bm{x}) & \text{otherwise}  
    \end{cases}       
\end{equation}
where, $q_1(\bm{x}) = e^{(-x_1^2 - x_2^2)} - x_1^3 - x_2^3$ and $q_x(\bm{x}) = 1 + q_1(\bm{x}) + \frac{1}{8}(x_1^2 + x_2^2)$.

\begin{figure*}[htbp]
  \centering
    \includegraphics[width=0.8\textwidth]{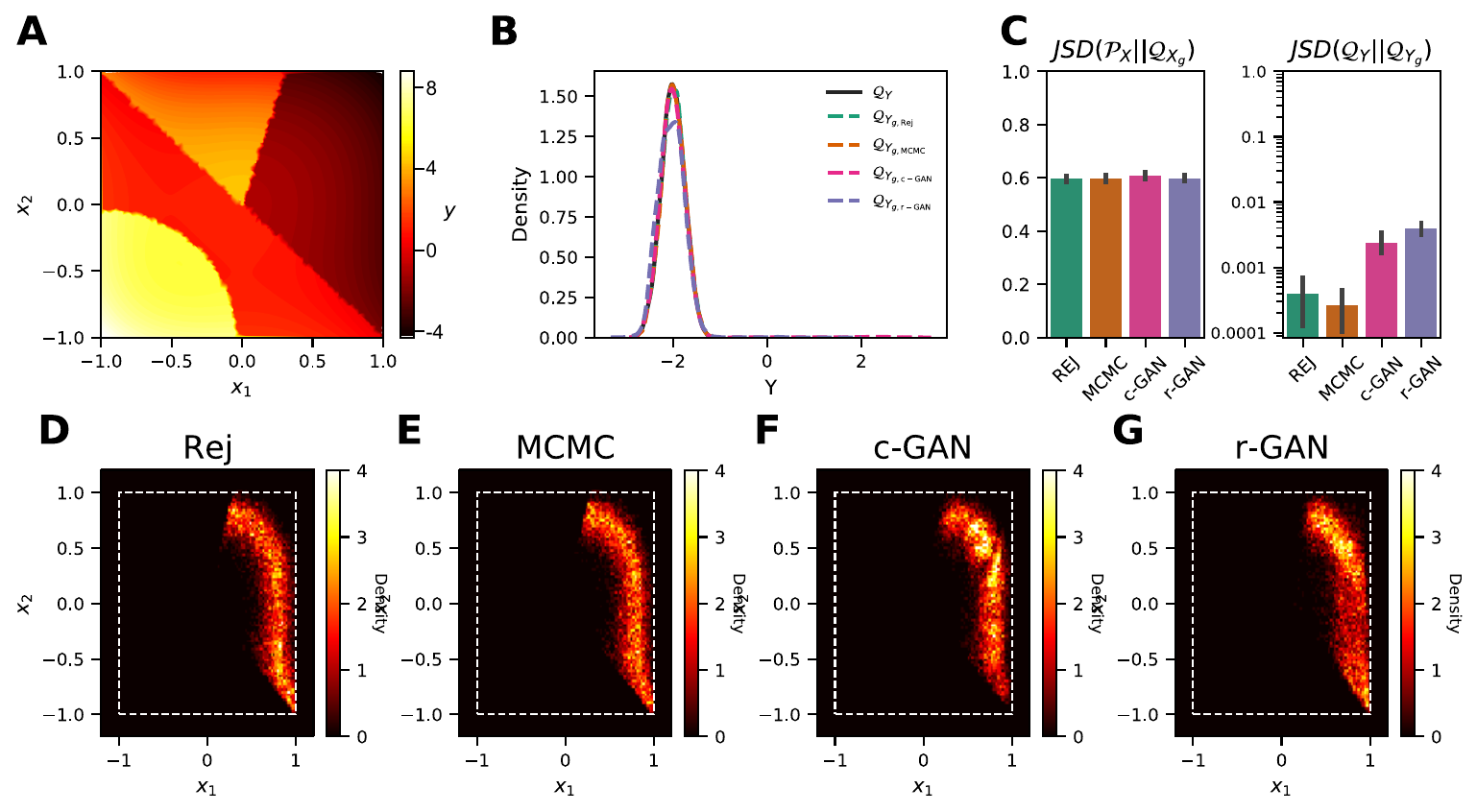}
\caption{Piecewise smooth test function. \textbf{A}. Heat map of $y$ over the prior. \textbf{B}. KDE of the desired target output distribution $\mathcal{Q}_Y$ and the generated (inferred) output distribution $\mathcal{Q}_{Y_g}$ using Rejection, MCMC, c-GAN and r-GAN (in green, orange, pink and purple, respectively). \textbf{C}. Left: estimated JS-divergence between samples from $\mathcal{P}_X$ and $\mathcal{Q}_{X}$ and Right: estimated JS-divergence between samples from $\mathcal{Q}_Y$ and $\mathcal{Q}_{Y_g}$ for all methods. \textbf{D-F}. 2D histograms of $\mathcal{Q}_X$ for Rejection (\textbf{D}), MCMC (\textbf{E}), c-GAN (\textbf{F}), and r-GAN (\textbf{G}) for $x_1$ and $x_2$. The dashed rectangle denotes the bounds set by the prior $\mathcal{P}_X$.}
  \label{fig:piecewise}
\end{figure*}

The distribution of observations $\mathcal{Q}_Y$ was $\mathcal{N}(-2.0, 0.25^2)$, and $\mathcal{P}_X$ for both $x_1$ and $x_2 \sim \mathcal{U}(-1, 1)$. Figure \ref{fig:piecewise}A shows a heat map of $y$ over $\mathcal{P_X}$ for \eqref{eq:piecewise}. 

In this example, we incorporated a surrogate model in the form of a feedforward network trained to approximate $\bf{y} = M(\bf{x})$ as the model node in the r-GAN. The surrogate network consisted of $4$ dense layers with $400$ nodes per layer, ReLU activation, and a dropout rate of $0.1$ between layers.

Calculating outputs according to \eqref{eq:piecewise} for the inferred input parameter samples $\mathcal{Q}_{X_g}$ generated by each method, shown in Figures \ref{fig:piecewise}D-G, resulted in model output distributions $\mathcal{Q}_{Y_g}$ via Rejection, MCMC, c-GAN and r-GAN, each of which matched the target observation density (Figure \ref{fig:piecewise}B). 

\subsection{An ordinary differential equation model}

\begin{figure*}[htbp]
  \centering
    \includegraphics[width=0.8\textwidth]{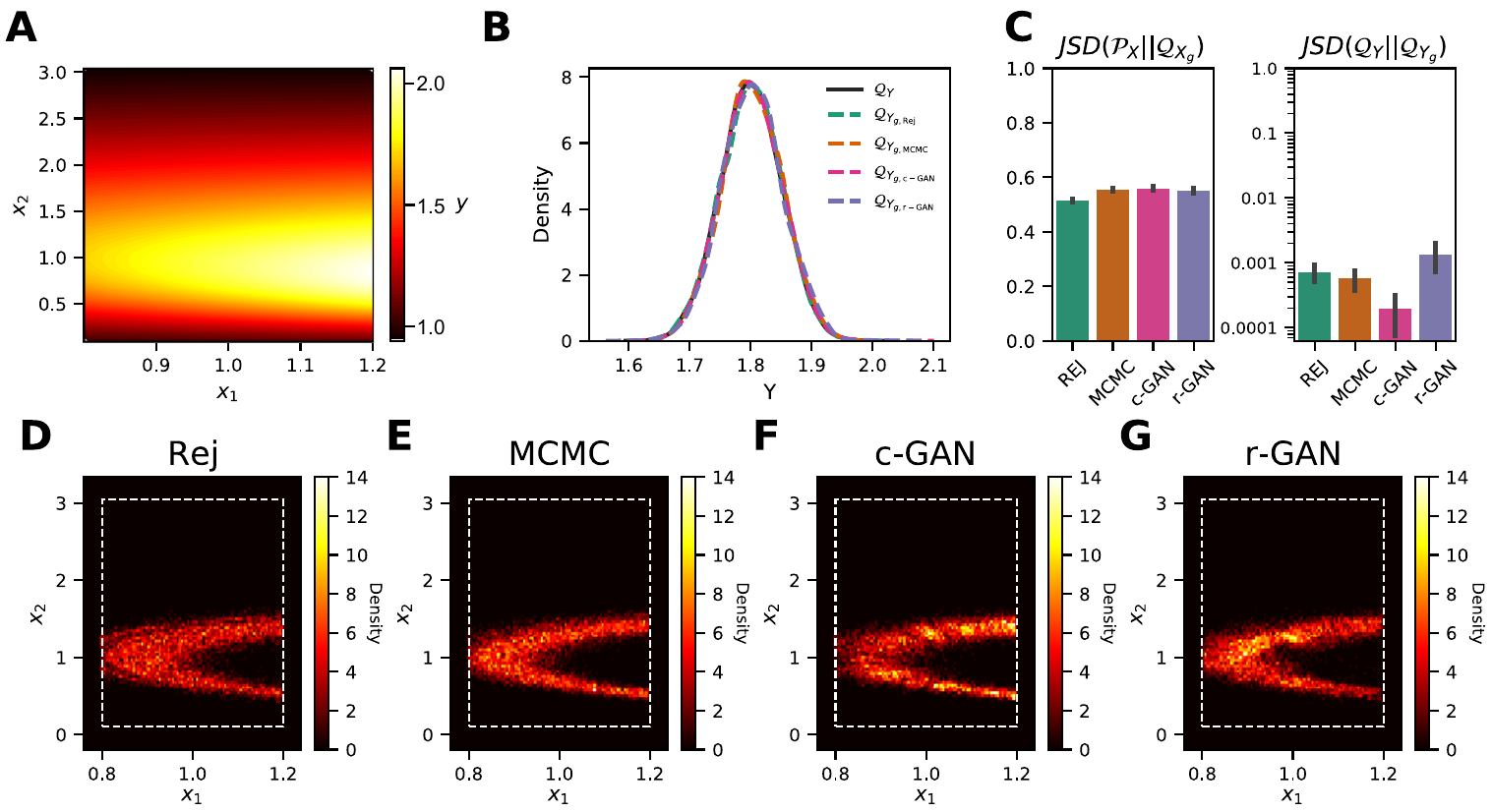}
\caption{ODE function test. \textbf{A}. Heat map of $y$ over the prior. \textbf{B}. KDE of the desired target output distribution $\mathcal{Q}_Y$ (black) and the generated (inferred) output distributions $\mathcal{Q}_{Y_g}$ using Rejection, MCMC, c-GAN and r-GAN (in green, orange, pink and purple, respectively). \textbf{C}. Left: estimated JS-divergence between samples from $\mathcal{P}_X$ and $\mathcal{Q}_{X}$ and Right: estimated JS-divergence between samples from $\mathcal{Q}_Y$ and $\mathcal{Q}_{Y_g}$ for all methods. \textbf{D-F}. 2D histograms of $\mathcal{Q}_X$ for Rejection (\textbf{D}), MCMC (\textbf{E}), c-GAN (\textbf{F}), and r-GAN (\textbf{G}). The dashed rectangle denotes the bounds set by the prior $\mathcal{P}_X$.}
  \label{fig:ode}
\end{figure*}

Finally, we tested the Rejection, MCMC, c-GAN and r-GAN methods on a model represented by an ordinary differential equation (ODE) with 2 input parameters $\bm{x}=(x_1, x_2)$,

\begin{equation}\label{eq:ode}
    \begin{split}
        \frac{df(t;x_1, x_2)}{dt} = & x_1 sin (x_2 f(t;x_1,x_2)), \quad \\ 
                                    & 0 < t < 2 \text{ and } f(t = 0) = 1\text{.}
    \end{split}
\end{equation}
The target feature of interest is $y(\bm{x}) = \frac{1}{2}\int_0^2 f(t; \bm{x}) dt$. Since the model is an ODE, we could have directly incorporated it into the deep learning networks for inference using a differentiable ODE solution \citep{chen2019neural}. Another option would be to solve the differential equation analytically and use the closed form solution in the deep learning network. However, as in the piecewise function example, we chose to build a forward model surrogate. To train the surrogate model, we sampled $10,000$ points from the prior and obtained the target feature of interest for all the training points by solving numerically the differential equation using the Python scipy module. Figure \ref{fig:ode}A shows the heat map of the feature of interest estimated via the trained surrogate forward model over the prior. 

The distribution of the target observation $\mathcal{Q}_Y$ was $\mathcal{N}(1.8, 0.05^2)$, and the input parameter prior distribution $\mathcal{P}_X$ with $x_1 \sim \mathcal{U}(0.8, 1.2)$ and $x_2 \sim \mathcal{U}(0.1, \pi - 0.1)$ as in \citep{breidt2011measure}. The joint distribution of the parameters $x_1$ and $x_2$ obtained using Rejection, MCMC c-GAN, r-GAN and MCMC is plotted in Figures \ref{fig:ode}D, E, F and G, respectively. Simulating the ODE model with the inferred input parameter samples produces distributions of outputs that match the target observation density accurately (Figure \ref{fig:ode}B).

\section{GAN configuration} 

\subsection{GAN stabilization and training}

\begin{figure*}[h!]
  \centering
    \includegraphics[width=0.7\textwidth]{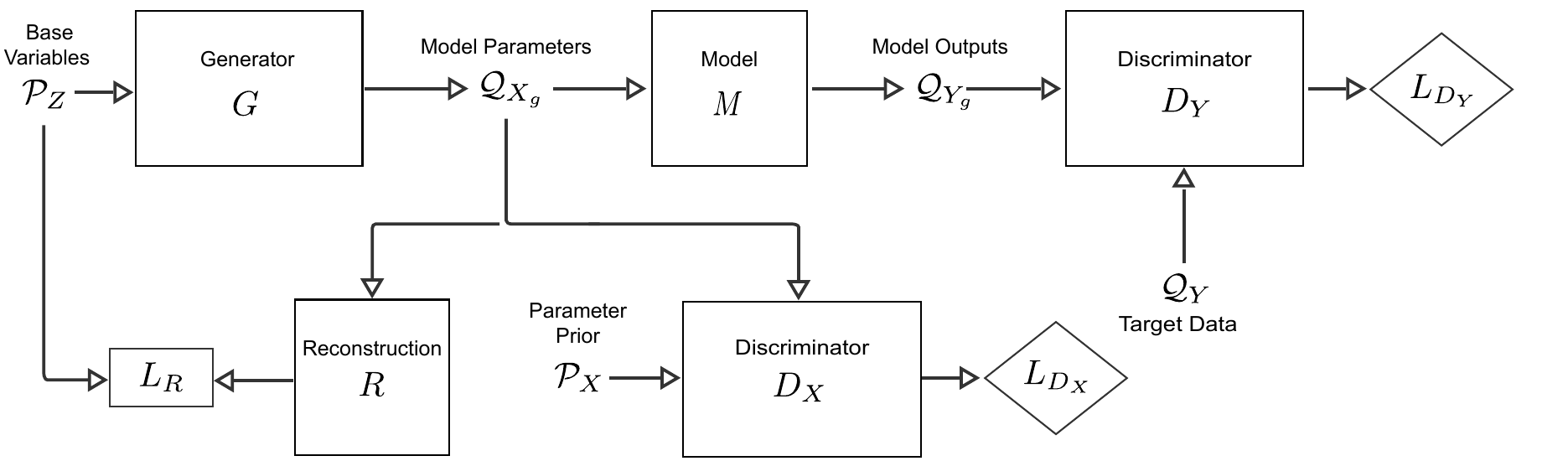}
  \caption{r-GAN architecture, showing reconstruction network $R$ and reconstruction loss $L_{R}$.}
  \label{fig:rgan_w_rec}
\end{figure*}

Reliably training GANs can be challenging due to known problems, such as mode collapse. Several approaches have aimed to stabilize adversarial networks \citep{wiatrak2019stabilizing}. However, in practice, it remains very difficult to find a robust and reliable method for GAN stabilization that works for a broad spectrum of generative models. One class of relatively simple stabilization algorithms is designed to increase the entropy of samples produced by the generator. This is accomplished by increasing mutual information between the input and output of the generator network, which in this case is equivalent to maximizing the entropy of the output. The loss function of the generator is therefore augmented by a mutual information term computed using dual representations of KL-divergence \citep{belghazi2018mine} or reconstruction networks as in VEEGAN \citep{srivastava2017veegan}. Here, we incorporate the reconstruction network approach, training the reconstruction network to reproduce the latent distribution $\mathcal{P}_{Z}$ from samples from the generator. We simplify the approach of VEEGAN, including only the $\ell_{2}$ component of the reconstructor loss, excluding the cross-entropy term and removing the dependence of the discriminator on $\bm{z}$. The r-GAN architecture diagram in the main manuscript excluded the reconstruction network used for stabilization for simplicity. The complete r-GAN, including reconstruction used for training all examples (except SIP extensions in section \ref{sec:interventionSIP}), is shown here in Figure \ref{fig:rgan_w_rec}.

\paragraph{r-GAN loss functions} In Figure \ref{fig:rgan_w_rec}, discriminator $D_{Y}$ distinguishes between samples from the distribution $\mathcal{Q}_{Y}$ and samples generated by the generator $G$ forwarded through the mechanistic model, for which the standard loss,

\vspace*{-0.9\baselineskip}
\begin{equation}
    \begin{split}
        L_{D_{Y}} = & \E_{y \sim \mathcal{Q}_{Y}} \text{log}[D_{Y}(y)] + \\
                    & \E_{z\sim \mathcal{P}_{Z}}, \text{log}[1-D_{Y}(M(G(z)))],
    \end{split}
\label{eq:LDY}
\end{equation}
is maximized. Discriminator $D_{X}$ distinguishes between samples from the prior over mechanistic parameters $\mathcal{P}_{X}$ and samples generated by $G$, for which the standard loss,

\vspace*{-0.9\baselineskip}
\begin{equation}
    \begin{split}
        L_{D_{X}} = & \E_{x \sim \mathcal{P}_{X}} \text{log}[D_{X}(x)] \, + \\
                    & \E_{z\sim \mathcal{P}_{Z}} \text{log}[1-D_{X}(G(z))],
    \end{split}
\label{eq:LDX}
\end{equation}
is maximized. The reconstruction network $R$ aims to reproduce the original base distribution $Z$ from samples generated by $G$, for which the squared loss is calculated according to

\vspace*{-0.9\baselineskip}
\begin{equation}
    \begin{split}
        L_{R} = \E_{z\sim \mathcal{P}_{Z}}||z - R(G(z))||^{2}\text{.}
    \end{split}
\label{eq:LR}
\end{equation}

The generator network $G$ generates mechanistic parameter sets from the base variable $Z$, for which losses are calculated from both $D_{Y}$ and $D_{X}$ according to

\vspace*{-0.9\baselineskip}
\begin{equation}
    \begin{split}
        L_{G_{Y}} = & \E_{z\sim \mathcal{P}_{Z}} -\text{log}[1-D_{Y}(M(G(z))] + \\
                    & \E_{z\sim \mathcal{P}_{Z}}, \text{log}[D_{Y}(M(G(z)))], \\
        L_{G_{X}} = & \E_{z\sim \mathcal{P}_{Z}} -\text{log}[1-D_{X}(G(z)] + \\
                    & \E_{z\sim \mathcal{P}_{Z}}, \text{log}[D_{X}(G(z))].
    \end{split}
\label{eq:LG2}
\end{equation}
The total loss for $G$ is then the weighted sum loss,

\vspace*{-0.9\baselineskip}
\begin{equation}
    \begin{split}
        L_{G} = w_{Y}L_{G_{Y}} + w_{X}L_{G_{X}} + w_{R}L_{R},
    \end{split}
\label{eq:LG}
\end{equation}
which is minimized, where $w_{Y} = 1.0$, $w_{X} = 0.1$, and $w_{R} = 1.0$ are default weights.

We used the Adam optimizer with step size of $0.0001$ for $G$ and $R$, and $0.00002$ for $D_{X}$, and $D_{Y}$. The $\beta_{1}$ and $\beta_{2}$ parameters of the Adam optimizer were set to default values of 0.9 and 0.999, respectively, as suggested in \citep{Kingma2015}, and mini-batch size was 100. Training was performed in two stages. First, $G$, $R$ and $D_{X}$ were trained together, with $w_{X} = 1.0$ and the $L_{D_{Y}}$ term removed in \eqref{eq:LG} (i.e., $w_{Y} = 0$) for 100 epochs to initialize $G$ by minimizing $D(\mathcal{P}_{X}||\mathcal{Q}_X{_{g}})$. Second, the full GAN was trained for 200 epochs on a dataset $y \sim \mathcal{Q}_{Y}$ comprising MNIST training images for the super-resolution imaging problem, or $10,000$ samples for the synthetic datasets used in the other experiments ($30,000$ samples for the high-dimensional Rosenbrock test in section \ref{sec:roshd}).

\subsection{GAN configuration for images}

\begin{figure}[htbp]
  \centering
    \includegraphics[width=1.\linewidth]{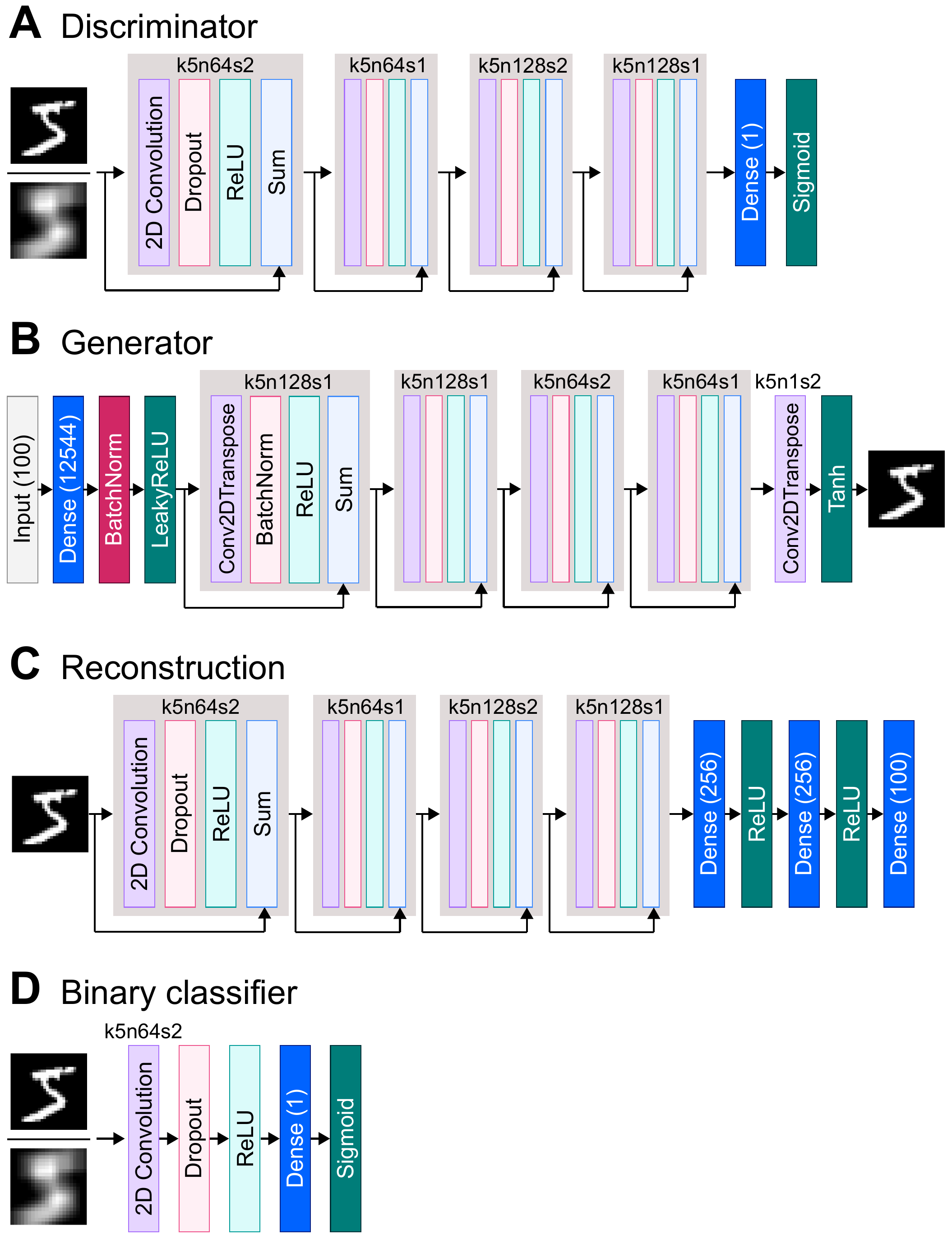}
  \caption{Network configurations used in GAN and r-GAN for super-resolution imaging. Grey boxes outline `residual blocks', with black arrows showing `skip connections', which used a $1\times1$ convolution when size changed between residual blocks' inputs and outputs. \textbf{A}. Network used for discriminators $D_{X}$ and $D_{Y}$, with either high-resolution (HR) or low-resolution (LR) images as inputs, respectively. \textbf{B}. Network structure used for generator $G$. Input derived from the base distribution $\mathcal{P}_{Z}$, and output formed a HR image. \textbf{C}. Network used for reconstruction network matches the discriminator network, but with additional final layers to recreate $\mathcal{P}_{Z}$. \textbf{D.} Network used for binary classifiers trained for JS-divergence calculations.}
  \label{fig:gan_for_images_diagram}
\end{figure}

The network configurations used for the discriminators, generator, and reconstruction network in the r-GAN for super-resolution imaging are shown in Figure \ref{fig:gan_for_images_diagram}. These networks were incorporated into the r-GAN architecture shown in Figure \ref{fig:rgan_w_rec}. The binary classifier network was used for JS-divergence calculations on samples from either the full MNIST dataset or subsets according to the labels, and samples from trained GAN or r-GAN generators (see section 6 of the main manuscript). Dropout rate was 0.1 in the discriminators, 0.4 in the reconstruction network, and 0.6 in the binary classifiers. Image size was $28\times28$ for high resolution (HR) images, and $22\times22$ for low resolution (LR) images.

\subsection{GAN configuration for test functions}

All test functions except the high-dimensional Rosenbrock had the same structure, with 2 inputs ($\bm{x}$) and 1 output ($\bm{y}$). We used identical c-GAN and r-GAN networks in all cases, with the generator, discriminator, and reconstruction networks consisting of densely connected, feedforward layers, and using either spectral normalization \citep{miyato2018spectral} (in $D_{Y}$ discriminators) or dropout (in $D_{X}$ discriminators) for regularization. Table \ref{tab:test_func_nets} details the configuration of each network. All test function examples used $w_{X} = 0.03$, $w_{R} = 3.0$ in the second stage of training, and Adam optimizer step sizes of $0.0001$ for $G$ and $R$ and $0.00001$ for $D_{X}$ and $D_{Y}$.

\begin{table}[htbp]
\caption{Neural networks used in c-GAN and r-GAN architectures.}
\label{tab:test_func_nets}
\vskip 0.1in
\begin{center}
\begin{small}
\begin{sc}
\begin{tabular}{p{0.12\columnwidth}p{0.12\columnwidth}p{0.09\columnwidth}p{0.16\columnwidth}p{0.11\columnwidth}p{0.16\columnwidth}}
Network & Hidden layers & Nodes per layer & Spectral \space\space\space\space\space Normalization & Dropout rate & Activation function \\
$D_{X}$    & 8 & 100 & No & 0.01 & ReLU \\
$D_{Y}$    & 8 & 100 & Yes & 0.00 & ReLU \\
$G$        & 8 & 100 & No & 0.0 & ReLU \\
$R$        & 8 & 100 & No & 0.0 & ReLU \\
\end{tabular}
\end{sc}
\end{small}
\end{center}
\vskip -0.1in
\end{table}

\subsection{GAN configuration for SIP extensions}

For the intervention SIP examples described below in section \ref{sec:interventionSIP} and shown in Figure \ref{fig:tgan}, we used networks with details shown in Table \ref{tab:tgan_nets}. In the shared parameter study, with multiple generators, the three sets of base variables $\mathcal{P}_{z1}$, $\mathcal{P}_{z2}$ and $\mathcal{P}_{z3}$ were concatenated as the target for the reconstruction network, and the reconstruction network took all generated $\bm{x}$ values $\mathcal{Q}_{X_{s,g}}$, $\mathcal{Q}_{X_{c,g}}$ and $\mathcal{Q}_{X_{d,g}}$ as input. The intervention examples were trained for $200$ epochs at both training stages. The shared parameter study used $w_{X} = 0.1$, $w_{R} = 3.0$, and Adam optimizer step sizes of $0.0001$ for $G$ and $R$, and $0.00002$ for $D_{X}$, and $D_{Y}$. The explicit map study used the same settings, except for $w_{X} = 0.03$, and Adam optimizer step sizes of $0.00001$ for $D_{X}$ and $D_{Y}$.

\begin{table}[htbp]
\caption{Neural networks used in intervention r-GAN architectures.}
\label{tab:tgan_nets}
\vskip 0.1in
\begin{center}
\begin{small}
\begin{sc}
\begin{tabular}{p{0.12\columnwidth}p{0.12\columnwidth}p{0.09\columnwidth}p{0.16\columnwidth}p{0.11\columnwidth}p{0.16\columnwidth}}
\toprule
Network & Hidden layers & Nodes per layer & Spectral \space\space\space\space\space Normalization & Dropout rate & Activation function \\
\midrule
Shared & & & & & \\
\midrule
$D_{X}$    & 8 & 80 & No & 0.01 & ReLU \\
$D_{Y}$    & 8 & 80 & Yes & 0.00 & ReLU \\
$G$        & 8 & 80 & No & 0.0 & ReLU \\
$R$        & 8 & 240 & No & 0.0 & ReLU \\
\midrule
Explicit & & & & & \\
\midrule
$D_{X}$    & 8 & 130 & No & 0.01 & ReLU \\
$D_{Y}$    & 8 & 130 & Yes & 0.00 & ReLU \\
$G$        & 8 & 80 & No & 0.0 & ReLU \\
$R$        & 8 & 130 & No & 0.0 & ReLU \\
\bottomrule
\end{tabular}
\end{sc}
\end{small}
\end{center}
\vskip -0.1in
\end{table}

\subsection{Computational resources}

Custom code for r-GAN was developed in TensorFlow and Pytorch, available at https://github.com/IBM/rgan-demo-pytorch. All experiments were conducted on a single NVIDIA V100 GPU. r-GAN training in the super-resolution imaging example took approximately $5$ minutes to compute. The test function examples each took approximately $1.5$ minutes to compute for MCMC, $2$ minutes for c-GAN training, and $5.5$ minutes to compute both stages of r-GAN training. The intervention examples took approximately $11$ minutes to compute both stages of r-GAN training.


\section{Classifiers for JS-divergence estimation and regularization of GAN discriminators}


We estimated JS-divergence between $\mathcal{P}_X$ and $\mathcal{Q}_{X_g}$, or between $\mathcal{Q}_Y$ and $\mathcal{Q}_{Y_g}$, using the density ratio trick with a classifier trained to distinguish between samples from the two distributions \citep{Sugiyama2012}. For samples $x_{1}$ and $x_{2}$ from two distributions $\mathcal{X}_{1}$ and $\mathcal{X}_{2}$, JS-divergence was calculated according to


\begin{equation}
    \begin{split}
        D_{JS_{S}}(\mathcal{X}_{1}||\mathcal{X}_{2}) & \approx \\
        & \frac{1}{2}\Bigg(\frac{1}{n_{1}}\sum_{i=1}^{n_{1}}\Big[\log_2(S(x_{1,i}))\Big]+ \\ 
        & \frac{1}{n_{2}}\sum_{j=1}^{n_{2}}\Big[\log_2(1-S(x_{2,j}))\Big]+2\Bigg),
    \end{split}
\label{eq:DJSS}
\end{equation}
where $S$ is a classifier trained to distinguish samples from $\mathcal{X}_{1}$ from samples from $\mathcal{X}_{2}$ (see Figure \ref{fig:gan_for_images_diagram}D), and $n_{1}$ and $n_{2}$ are the number of samples in $x_{1}$ and $x_{2}$. 

In practice, for each test function and for each method (Rejection, MCMC, c-GAN and r-GAN), we generated $10,000$ samples from both the target distribution and inferred distribution, and further sampled randomly $1,000$ of each of those samples as test sets for the JS-divergence calculation using \eqref{eq:DJSS}. With the remaining $9,000$ samples, we trained $5$ classifiers using a different random subset of $7,200$ samples as the training set and $1,800$ samples as the validation set for each. The error bars in all JSD measures in the figures show the standard deviation of JSD values across these $5$ different trained classifiers. The classifier was a 2-layer dense network with $100$ nodes per layer and softplus activation, and a single-node output layer, trained with the binary cross-entropy loss between the two sets of samples for $1,000$ epochs with batch size $1,000$, using early stopping with patience of $40$ epochs based on the loss calculated on the validation set.

As we mentioned in section 4.1 of the main manuscript, estimating divergence measures in a high-dimensional space such as images is extremely challenging. In such a high-dimensional space, convolutional classifiers become very effective at discriminating between ``true'' and generated samples. We used a simple network, relative to the discriminator networks in the r-GAN used for super-resolution imaging, for the binary classification of images shown in Figure \ref{fig:gan_for_images_diagram}D. This classification network was simple and highly regularized (dropout rate of $0.6$) in order to prevent perfect classification of samples, but the nature of the regularization of the classifier played an important role in the JS-divergence values that are estimated. For JS-divergence estimation in the super-resolution imaging example (see Table I in the main manuscript), we trained classifiers $10$ times for each comparison, and report the mean JS-divergence estimate for each. Standard deviations were below 2 decimal places, so were not reported.

The choice of discriminator regularization plays an important role in the solutions to SIP found by r-GAN, especially for high-dimensional problems. Because the generator defines an implicit density model that is constructed during training, the extent to which the discriminator is able distinguish real observations from generated samples strongly influences the parameter distributions that are generated by the final trained model. An extremely powerful discriminator should lead to generation of samples that precisely match the observed samples, whereas a less powerful, more highly regularized discriminator should lead to generation of samples that are less precisely aligned to the observed samples. In the super-resolution imaging example, this manifests as a larger JS-divergence between '5's in the MNIST dataset and samples generated by the r-GAN (0.22) than between '5's in the MNIST dataset and samples generated by the GAN with PULSE model (0.07), which finds exactly one output sample per target image.

An important avenue for future work is to establish methods for selecting among reasonable options for discriminator regularization for a particular problem. This could be accomplished, for example, by training classifiers with the configuration of a discriminator testing different regularization methods on the target dataset using methods analogous to cross-validation. 

\nolinenumbers
\bibliographystyle{IEEEtran}
\bibliography{supplement}
